\documentclass[runningheads]{llncs}
\usepackage[T1]{fontenc}
\usepackage{graphicx}
\usepackage{booktabs}
\usepackage[misc]{ifsym}
\newcommand{\corr}{(\Letter)}
\usepackage{mwe}
\usepackage{amssymb}
\usepackage{stmaryrd}
\usepackage{amsmath,amsfonts}
\usepackage{array}
\usepackage{textcomp}
\usepackage{stfloats}
\usepackage{url}
\usepackage{verbatim}
\usepackage{graphicx}
\usepackage{cite}
\usepackage{xcolor}
\usepackage{multirow}
\usepackage{amsmath,amssymb,amsfonts}%
\usepackage{stmaryrd}
\usepackage{rotating}
\usepackage{bm}
\usepackage{hyperref}
\usepackage{booktabs}
\usepackage{subcaption}%
\usepackage{listings}%

\usepackage[ruled,vlined,linesnumbered]{algorithm2e}

\usepackage{todonotes}
\usepackage{comment}
 
\usepackage{color}
\definecolor{Orange}{rgb}{1,0.5,0}
\definecolor{Red}{rgb}{1,0,0}
\definecolor{Blue}{rgb}{0,0,1}

\begin{document}

\title{Addressing Imbalance in Multi-Label Data via Label-Specific Distance-based Oversampling}

\titlerunning{Addressing Imbalance in Multi-Label Data via LSD-based Oversampling}

\author{Bin Liu\inst{1} \and
Jun Wu\inst{1,2} \and
Haoyu Peng\inst{1,2} \and
Ao Zhou\inst{3} \corr \and
Jin Wang\inst{1} \and
QiaoSong Chen\inst{1} \and
Grigorios Tsoumakas\inst{4}
}


\authorrunning{B. Liu et al.}


\institute{Key Laboratory of Data Engineering and Visual Computing, Chongqing University of Posts and Telecommunications, China \email{\{liubin,wangjin,chenqs\}@cqupt.edu.cn}
\and
School of Computer Science and Technology, Chongqing University of Posts and Telecommunications, China
\email{\{s240231056,s240231185\}@stu.cqupt.edu.cn}
\and
State Key Laboratory of Novel Software Technology, Nanjing University, China \email{zacqupt@gmail.com}
\and
School of Informatics, Aristotle University of Thessaloniki, Greece 
\email{greg@csd.auth.gr}
}




\maketitle              

\begin{abstract}
The complex imbalanced label distribution poses a crucial challenge to multi-label classification, as most classifiers are biased towards the majority class and high-frequent labels. Oversampling is an efficient and flexible solution that augments instances to provide a more balanced training dataset for multi-label classifiers. Most existing oversampling methods create synthetic instances in a heuristic way that essentially relies on neighborhood information retrieved using Euclidean distance within the entire feature space. However, they fail to consider the varying semantic relevance of features to different labels, leading to label inconsistency among proximate neighbors and further introducing label confusion and overfitting to synthetic instances. To overcome the above issue, we propose a novel sampling approach called Label-Specific Distance-based Multi-Label Oversampling (LSDMLO) that creates more useful and well-labeled synthetic instances to address the imbalance in multi-label datasets. LSDMLO derives the label-specific distance to identify label-consistent neighbors based on the weighted pertinent feature space, which facilitates selecting seed instances that express more label correlations in boundary areas and generating synthetic instances aligned with the label distribution of original data. The comprehensive experiments verify that the proposed LSDMLO outperforms the state-of-the-art multi-label sampling approaches under various base classifiers.

\end{abstract}

\section{Introduction}
In Multi-Label Classification (MLC), an instance can be associated with multiple labels simultaneously, allowing for richer semantics to be expressed by each data point~\cite{mllreview1}. MLC has been widely applied in diverse domains, such as text categorization~\cite{text2}, image classification~\cite{image1}, and video annotation~\cite{video1}.
A key challenge in MLC is the inherent imbalance of multi-label datasets \cite{review}. 
For each label, its relevant instances (tagged as ``1'') are typically much fewer than its irrelevant ones (tagged as ``0''), leading to imbalance within labels. In addition, label frequencies (the number of ``1''s in each label) might vary widely, causing imbalance between labels. Apart from imbalance at the global level, local imbalance, which represents the label distribution within neighborhoods, is crucial to assess the difficulty of learning a multi-label classifier on the dataset \cite{MLSOL}. Most multi-label classifiers are biased towards the majority class (``0'') and high-frequency labels due to the complex nature of imbalance in multi-label data \cite{MLKNN,ECC}.

There are two types of solutions to deal with imbalanced multi-label data, namely algorithm adaptation methods and sampling methods \cite{review}.  
Algorithm adaptation methods extend existing multi-label classifiers \cite{ECCRU3,costsvm} or design new algorithms \cite{COCOA,SOSHF,MKML} that are resilient to the imbalance of multi-label datasets. 
Multi-label classification is more intricate than single-label (binary or multi-class) classification tasks, so there is no single algorithm adaptation method that can fit all situations \cite{generalization}. 
Sampling approaches mitigate imbalance in multi-label data via instance augmentation or removal, leading to a more balanced dataset for training a multi-label classifier \cite{ML,MLSMOTE,MLSOL,MLBOTE, MLONC, DR-SMOTE,AEMLO}.
Compared to algorithm adaptation methods, sampling approaches can be paired with any multi-label classifier, even an algorithm adaptation method. This flexibility makes them more appropriate for tackling diverse scenarios.

Most multi-label sampling methods rely on neighborhoods to guide the generation of new instances \cite{MLSMOTE, MLSOL, MLBOTE, MLONC, DR-SMOTE}, typically identified using Euclidean distance over the entire feature space. 
However, each label in multi-label data expresses different semantic meanings relevant to specific partial features \cite{LLSF}. For example, for tagging a document with the \textit{sports} label, features such as \textit{athlete names} and \textit{event locations} are more important than features like \textit{technical terminologies} and\textit{ company names}, which are more relevant to the \textit{technology} label. 
The traditional Euclidean distance-based neighbor retrieval does not distinguish the importance of features for different labels, failing to capture the most relevant instances for a specific label. This may lead to label inconsistency among nearest neighbors, which limits the quality of synthetic instances generated by multi-label sampling methods. First, it is more challenging to determine the labels of a synthetic instance created by two instances associated with different labels. Existing methods usually handle this label discrepancy by adopting voting-based strategies~\cite{MLSMOTE} or distance-based labeling schemes~\cite{MLSOL}.
Second, relying on a small set of same-label neighbors for instance generation may increase the risk of overfitting. For example, MLBOTE~\cite{MLBOTE} requires the seed and reference instances to share the same labelset, which may leave very few candidates for rare label combinations and thus reduce the diversity of generated samples.



Apart from neighbor-based oversampling methods, AEMLO \cite{AEMLO} employs an autoencoder-based architecture with a tailored objective to generate synthetic instances in a latent feature space. However, it learns a global feature representation without considering label-specific feature relevance, which may mix features associated with different labels and result in label-inconsistent samples that fail to capture the discriminative characteristics of minority labels.

This paper proposes a Label-Specific Distance-based Multi-Label Oversampling method (LSDMLO) that creates more realistic and effective synthetic instances to address the imbalance in multi-label data.
To overcome the limitation of traditional Euclidean distance, we first propose the label-specific distance, which relies on the most pertinent feature subsets to identify neighbors assigned to the same label. The label-specific distance is derived based on feature importance for each label obtained by optimizing a convex problem that incorporates $\ell_1$ norm to sparsify feature coefficients and graph regularization for mining label correlations. 
Based on the label-specific distance, LSDMLO assesses candidate seed instances by considering the instance proximity to each label boundary as well as the extracted global and local label correlations.
Furthermore, label-specific distance-aware synthetic instance generation in LSDMLO guarantees the label consistency of synthetic instances in the important feature subspace. 
Experimental results on several multi-label datasets with various multi-label classifiers demonstrate the superiority of LSDMLO over state-of-the-art multi-label sampling methods. 


\section{Related Work}

\subsection{Multi-Label Classification}

Exploiting co-occurrence or exclusion relations among labels is crucial in MLC, and approaches are typically categorized by the order of label correlations they consider \cite{mllreview1,mllreview2}. 
\textit{First-order methods} ignore label correlations, treating each label independently. Binary Relevance (BR) decomposes the problem into multiple binary classification tasks \cite{BR,new_BR}, while Multi-Label k-Nearest Neighbors (MLkNN) predicts each label based on its posterior probability within local neighborhoods \cite{MLKNN,new_MLkNN}. 
\textit{Second-order methods} capture pairwise label correlations, such as Calibrated Label Ranking (CLR), which transforms MLC into a label ranking problem by fitting classifiers for label pairs \cite{CLR}. 
\textit{High-order methods} model dependencies among more than two labels. Classifier Chains (CC) construct a sequence of binary classifiers where each leverages predictions of preceding labels \cite{ECC,new_cc}. Random k-Labelsets (RAkEL) encodes random label subsets as multi-class tasks \cite{RAkEL}. 
Deep learning models such as C2AE~\cite{C2AE} and CLIF~\cite{CLIF}, which jointly learn feature and label embeddings or label-specific feature representations with objectives that account for label correlations.
Furthermore, recent studies have investigated batch selection strategies that prioritize instances according to hardness~\cite{zhou2024multi} or uncertainty~\cite{zhou2025batch}, thereby improving predictive performance and accelerating the convergence of deep multi-label models.

\subsection{Handling Class Imbalance in Multi-Label Data}
Dealing with the class imbalance in MLC can be divided into two main categories: algorithm adaptation and sampling methods. 

Algorithm adaptation modifies existing classifiers or designs new models to handle imbalanced labels. ECCRU~\cite{ECCRU3} extends ECC (ensemble of CCs) by combining undersampling with better use of majority instances, while ECC++~\cite{ECC++} integrates sampling, cost-sensitive learning, and threshold moving. COCOA~\cite{COCOA} builds multi-class classifiers for label pairs to mitigate imbalance and capture label correlations. 
BalanceMix~\cite{BalanceMix}, a deep learning model, augments low-confidence instances and adjusts label-specific losses to handle imbalance and noise.

Data sampling methods aim to obtain a more balanced training set during the data pre-processing stage.  
Unlike undersampling methods (e.g., MLRUS \cite{ML}, MLTL \cite{MLTL}) that may discard informative instances, oversampling approaches demonstrate superior performance  \cite{MLONC}.
MLSMOTE adapts SMOTE for multi-label data,
it creates synthetic instances near existing instances associated with minority labels and assigns labels to generated instances according to the local label statistics. 
MLSOL \cite{MLSOL} further considers local label imbalance and employs weight vectors and type matrices for seed instance selection and synthetic instance generation. 
MLBOTE \cite{MLBOTE} divides instances into self-boundary, cross-boundary, and internal categories, and determines specific instance weights and sampling rates for each group.
DR-SMOTE \cite{DR-SMOTE} generates diverse, reliable multi-label samples by using non-noisy seeds and reference samples to identify synthesis regions, creating new instances preserving the seed's labels.
MLONC \cite{MLONC} adaptively detects natural neighbors for each label and leverages label correlations to generate synthetic instances near decision boundaries. 
REMEDIAL decouples minority and majority labels by modifying the label space and copying instance features to avoid co-occurrence interference, and can be combined with oversampling methods like MLSMOTE to form RHwRSMT~\cite{REMEDIAL-HwR}. 
AEMLO \cite{AEMLO} is an autoencoder-based multi-label oversampling method that jointly embeds features and labels into a shared latent space, then generates synthetic instances non-linearly from minority samples while preserving label consistency. 



\section{Proposed Method}
\begin{figure*}[t]
    \centering
    \includegraphics[width=1.0\textwidth]{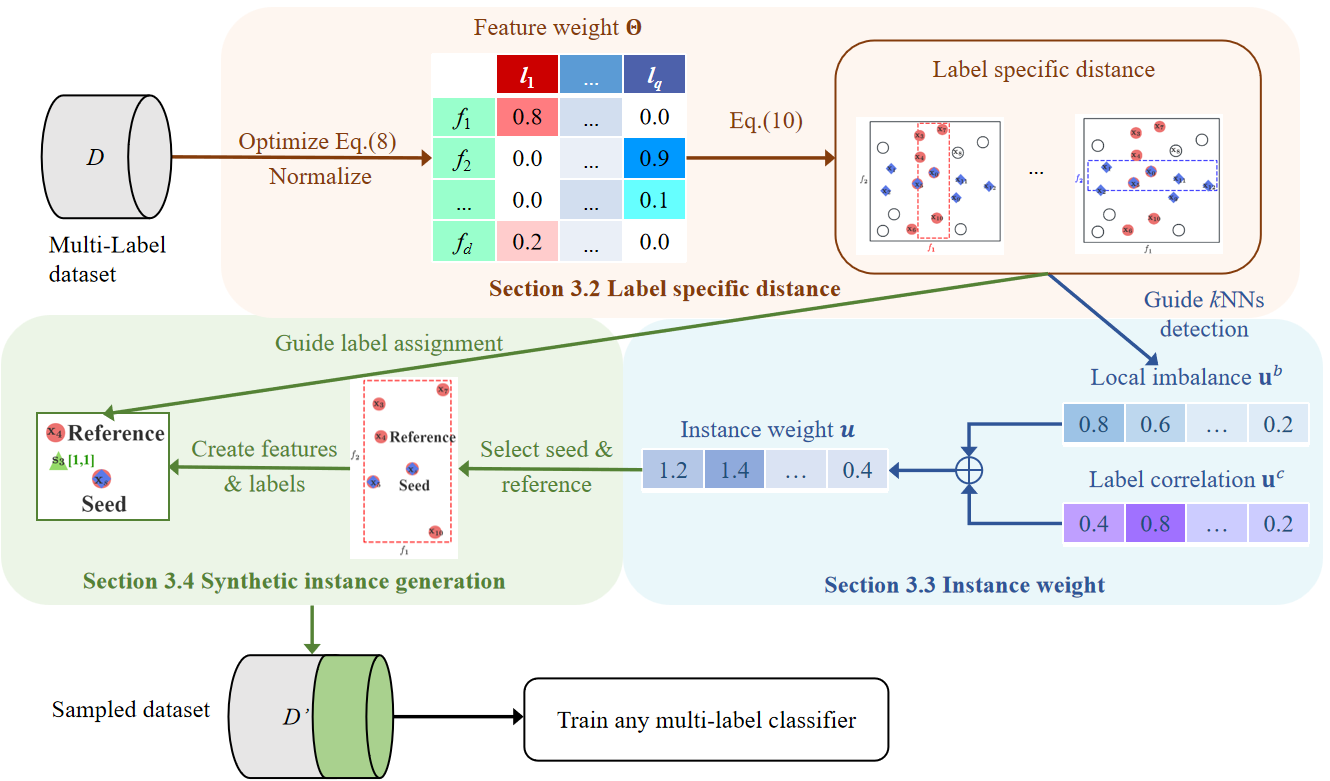}
    \caption{The workflow of LSDMLO}
    \label{fig:process}
\end{figure*}

LSDMLO addresses the imbalance in multi-label data by leveraging a label-specific distance measure to guide the synthetic instance augmentation process.
The workflow of LSDMLO is shown in Figure \ref{fig:process}.
and Algorithm \ref{al:LSDMLO}. 
Firstly, LSDMLO learns sparse label-specific feature weights and derives label-specific distances (Section \ref{LSD}).
Subsequently, the importance of instances is evaluated via the integration of local label imbalance and label correlations with the guidance of label-specific distances (Section \ref{WC}).
Finally, LSDMLO creates diverse and accurate synthetic instances using the label-specific distance-induced label assignment (Section \ref{IG}).

\begin{algorithm}[t]
\caption{LSDMLO}
\label{al:LSDMLO}
\SetKwData{Left}{left}\SetKwData{This}{this}\SetKwData{Up}{up}
\SetKwFunction{InitTypes}{InitTypes}\SetKwFunction{CreateIns}{CreateIns}
\SetKwInOut{Input}{input}\SetKwInOut{Output}{output}
\Input{multi-label dataset $D=\{\mathbf{X},\mathbf{Y}\}$, trade-off parameter $\lambda_{1}$ and $\lambda_{2}$, sampling ratio $p$}
\Output{sampled data set $D'$}
Feature weight $\mathbf{W}$ $\leftarrow \text{FeatureWOpti}(\mathbf{X}, \mathbf{Y}, \lambda_{1}, \lambda_{2})$  \tcc*[r]{Appendix Algorithm 1}
Get label-specific distance and $\mathbf{\Theta}$ \;
Instance weight $\mathbf{u}$ $\leftarrow \text{InsWeightCal}(D, \mathbf{A}, \mathbf{\Theta} )$ \ \tcc*[r]{Algorithm \ref{al:Weight_Calculate}}
Sampled dataset $D'$ $\leftarrow \text{InsGen}(D,\mathbf{u},L_{min},p)$ \tcc*[r]{Algorithm \ref{al:InstanceGeneration}}
\KwRet{$D'$} 
\end{algorithm}

\subsection{Notations}

Let $D=\{(\mathbf{x}_i,\mathbf{y}_i)\}_{i=1}^{n}$ be a multi-label dataset with $n$ instances, where $\mathbf{x}_i\in\mathbb{R}^{d}$ is a $d$-dimensional feature vector and $\mathbf{y}_i\in\{0,1\}^{q}$ is the corresponding label vector over the label set $L=\{l_1,\ldots,l_q\}$, where $y_{ij}=1$ indicates that $\mathbf{x}_i$ is relevant to label $l_j$, and $y_{ij}=0$ otherwise. The dataset can also be represented by a feature matrix $\mathbf{X}\in\mathbb{R}^{n\times d}$ and a label matrix $\mathbf{Y}\in\{0,1\}^{n\times q}$, whose $i$-th rows correspond to $\mathbf{x}_i$ and $\mathbf{y}_i$, respectively. The goal of multi-label classification is to learn a mapping $h:\mathbf{x}\rightarrow\{0,1\}^{q}$.

To characterize label imbalance in multi-label datasets, the Imbalance Ratio per Label ($IRLbl$) \cite{ML} is defined as
\begin{equation}
IRLbl_j = \frac{n_{\max}}{n_j}, \quad j=1,\ldots,q,
\end{equation}
where $n_j=\sum_{i=1}^{n} y_{ij}$ is the number of instances associated with label $l_j$, and $n_{\max}=\max_{1\le j \le q} n_j$. Label presence (1) and absence (0) are treated as the minority and majority classes, respectively. The Mean Imbalance Ratio ($MeanIR$) \cite{ML} averages IRLbl across all labels:
\begin{equation}
MeanIR = \frac{1}{q} \sum_{j=1}^{q} IRLbl_j,
\end{equation}
with larger values indicating stronger global label imbalance. A label $l_j$ is considered a \textit{minority label} if $IRLbl_j > MeanIR$, and the set of all minority labels is denoted by $L_{\text{min}}$. 
The Local Label Imbalance ($LImb$)~\cite{MLSOL} measures label disagreement within each instance's $k$-nearest neighbors ($k$NNs):
\begin{equation}
LImb = \frac{1}{q} \sum_{j=1}^{q} \frac{\sum_{i=1}^{n} b_{ij} y_{ij}}{\sum_{i=1}^{n} y_{ij}}, \quad
b_{ij} = \frac{1}{k} \sum_{(\mathbf{x}_m,\mathbf{y}_m)\in \mathcal{K}_{\mathbf{x}_i}} \llbracket y_{ij} \neq y_{mj} \rrbracket,
\label{LImb}
\end{equation}
where $\llbracket \pi \rrbracket$ is the indicator function and $\mathcal{K}_{\mathbf{x}_i}$ denotes the $k$NNs of $\mathbf{x}_i$ based on Euclidean distance. A larger $LImb$ indicates a more complex local label distribution, making the dataset harder to learn.

\subsection{Label-Specific Distance \label{LSD}}

Learning Label-Specific Features (LLSF) aims to derive the most pertinent and discriminative features for each label from the whole feature space, getting rid of the impact of irrelevant features and improving the performance of multi-label classifiers \cite{LLSF,LDL,CLML}. LLSF methods typically train a linear model with \textbf{$\ell_{1}$} norm regularization to obtain sparse feature weights for each label, i.e., only a small number of relevant features contribute to the prediction of each label.
Inspired by LLSF, we propose the following optimization problem to learn label-specific feature weights:
\begin{equation}
\label{1}
\mathop{min}\limits_{\mathbf{W}}\frac{1}{2}\left \| \mathbf{XW}-\mathbf{Y}\right \|_{F}^{2}+\lambda _{1} \left \| \mathbf{W}\right \|_{1} +\frac{\lambda _{2}}{2}tr(\mathbf{XW}\mathbf{\Lambda}(\mathbf{XW})^{\top})
\end{equation}
where $\mathbf{W}=[\mathbf{w}_{\cdot 1}, \mathbf{w}_{\cdot 2}, ..., \mathbf{w}_{\cdot q}] \in \mathbb{R}^{d \times q}$ denotes the coefficient matrix, and $\textbf{w}_{\cdot j}$ denotes the feature importance vector of label $l_j$. Specifically, $w_{ij}$ quantifies the strength of correlation between the $i$-th feature and label $l_j$.
In Eq.\eqref{1}, the first term minimizes the squared loss of predictions $\mathbf{XW}$ and truth labels $\mathbf{Y}$. 
The second term corresponds to the \textbf{$\ell_{1}$} norm regularization of $\mathbf{W}$, which enforces the sparsity on $\textbf{W}$. 
The third term applies graph regularization on model predictions to capture label correlations and implicitly mine the complex relationship between features and labels. $\mathbf{\Lambda}$ = $\mathbf{A}'-\mathbf{A}$ denotes the left-marginalized Laplacian matrix, $\mathbf{A} \in \mathbb{R}^{q \times q}$ represents label correlations with $A_{ij}$ calculated as the cosine similarity between the $i$-th and $j$-th column of $\mathbf{Y}$, $\mathbf{A}'$ is a diagonal matrix with ${A}'_{ii}=\sum_{j=1}^{q}A_{ij}$. 


The minimization of objective \eqref{1} is convex but non-smooth due to the $\ell_1$ norm term. Therefore, the Accelerated Proximal Gradient (APG) algorithm \cite{APG} is employed to obtain the optimal $\mathbf{W}$. The optimization details and the Lipschitz continuity proof are provided in \textit{Appendix Section~A1.1}.
It should be noted that Eq.\eqref{1} only includes basic $\ell_1$ norm and graph regularization to derive feature weights due to their simplicity and efficiency. 
More complex regularization terms, such as learning common features \cite{LDL} and capturing instance correlations \cite{CLML}, could also be used with our method.

The optimized $\mathbf{W}$ distinguishes pertinent and irrelevant features for each label, assessing the importance of discriminative features for each label. Therefore, we employ $\mathbf{W}$ to derive label-specific distances. Let $\mathbf{\Theta}=[\bm{\theta}_{\cdot 1}, \bm{\theta}_{\cdot 2}, ..., \bm{\theta}_{\cdot q}] \in \mathbb{R}^{d \times q}$ be the label-specific feature weight matrix, where $\bm{\theta}_{\cdot j} =[\theta_{1j}, \theta_{2j}, \cdots, \theta_{dj}]\in \mathbb{R}^{d}$ is the label-specific feature weight vector of $l_{j}$ via column-wise normalizing non-zero elements of $\mathbf{W}$:
\begin{equation}
\begin{split}
{\theta}_{ij}=\left\{\begin{matrix}
 \frac{\exp \left(\left | w_{ij} \right|\right)}{\sum_{i=1}^{d}\exp \left(\left | w_{ij} \right|\right) }  &  \text{if } w_{ij} \neq 0 \\
  0 & \text{otherwise}
\end{matrix}\right. 
\end{split}
\label{weight}
\end{equation}
Based on $\mathbf{\Theta}$, we can measure the distance between instances in label-specific ways, e.g., 
the $l_{j}$-specific distance between instance $(\mathbf{x}_{1},\mathbf{y}_{1})$ and $(\mathbf{x}_{2},\mathbf{y}_{2})$ is:

\begin{equation}
d(\mathbf{x}_{1}, \mathbf{x}_{2}, \bm{\theta}_{\cdot j})= \Big ( \sum_{i=1}^{d}  \theta_{ij} \left \|  \mathbf{x}_{1i}-  \mathbf{x}_{2i}\right \|^{2} \Big )^{\frac{1}{2}}  
\label{distance}
\end{equation}
\textit{Appendix Figure~A1} illustrates that label-specific distance retrieves more label-consistent neighbors than Euclidean distance, demonstrating its superior discriminative capability.

\subsection{Instance Weight \label{WC}}



\begin{figure*}[b]
  \centering
    \includegraphics[width=0.6\textwidth]{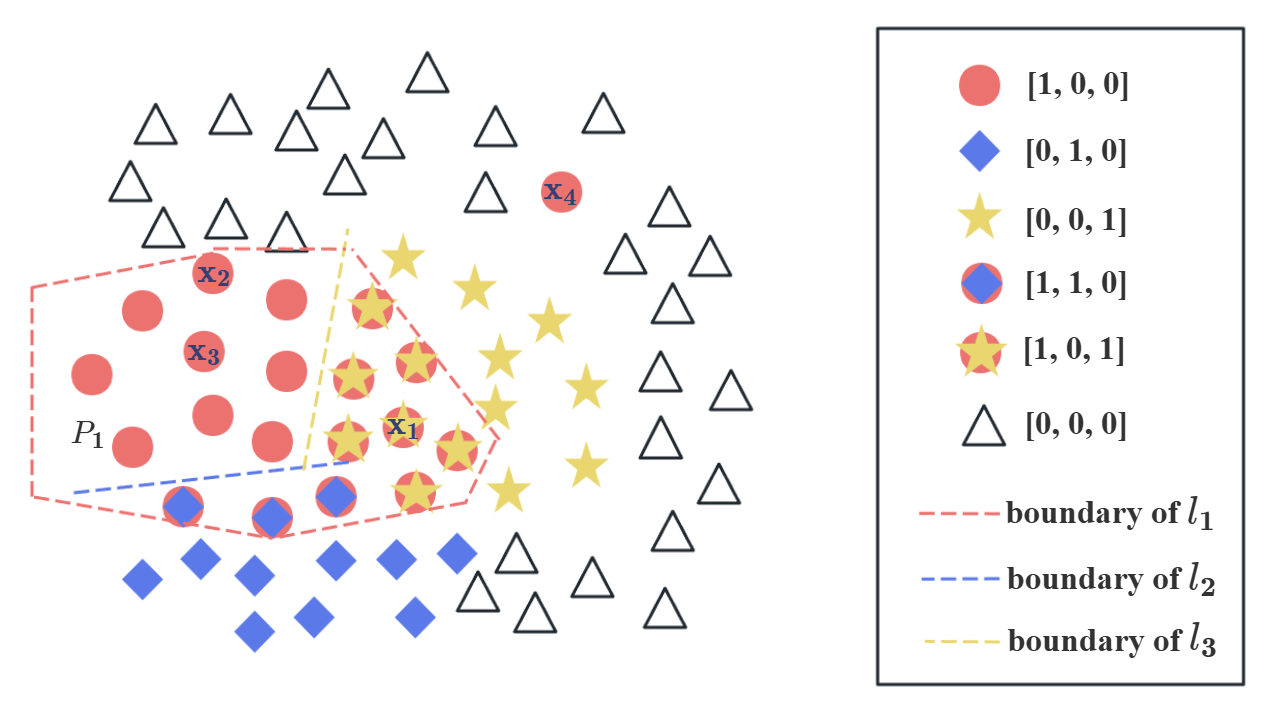} 
    \caption{A toy multi-label dataset that illustrates label boundaries, where $\mathbf{x}_{1}$ and $\mathbf{x}_{2}$ are borderlines that are near the boundary of $l_1$, $\mathbf{x}_{3}$ is an internal example positioned inside $P_1$ cluster, and $\mathbf{x}_{4}$ is an outlier.}
    \label{fig:boundary}
\end{figure*}

Instance weight is a crucial component of sampling methods, as it determines where synthetic instances are generated. 
Let ${P}_{j}=\left \{\left ( {\mathbf{x}}_{i},\mathbf{{y}}_{i} \right ) \mid {y_{ij}}=1\right \}$ be the positive training instance sets of label $l_j$. In the example toy dataset (Figure \ref{fig:boundary}), $P_1=\left \{ (\mathbf{x}_{1},\mathbf{y}_{1}),(\mathbf{x}_{2},\mathbf{y}_{2}),(\mathbf{x}_{3},\mathbf{y}_{3}),(\mathbf{x}_{4},\mathbf{y}_{4})\right \}$ is the set of instances associated with $l_1$. 
When considering each label boundary individually, instances close to the label boundary ($\mathbf{x}_{1}$ and $\mathbf{x}_{2}$) suffer more imbalanced label distribution (neighbor samples contain more inconsistent labels) than internal ones ($\mathbf{x}_{3}$), playing a more crucial role in the sampling process. 
Moreover, labels sharing strong dependencies usually co-exist in the same region, leading to overlapping multiple label boundaries, such as $l_1$ and $l_3$ associated with examples marked by a yellow star on a red circle. Instance $\mathbf{x}_{1}$, located in the overlapping boundary region of both $l_1$ and $l_3$, captures richer label correlations than $\mathbf{x}_{2}$, which is relevant to one label only. Creating a synthetic instance from $\mathbf{x}_{1}$ could introduce a positive instance for both $l_1$ and $l_2$, alleviating the imbalance of two labels. 
Therefore, LSDMLO assumes that the importance of a multi-label instance is influenced by two factors: (1) proximity to label boundaries, and (2) the number of minority labels associated with the instance.

The closeness of an instance to each label boundary could be assessed by local label imbalance \cite{MLSOL}. In order to evaluate the importance of the features for each label, we employ the label-specific distance to define an instance's local region for each label. 
Specifically, the label-specific distance-based local imbalance of $\mathbf{x}_i$ w.r.t. a minority label $l_j \in  L_{min}$ is defined as:
\begin{equation}
\hat{b}_{ij}=\left\{\begin{matrix}
\frac{1}{k}\sum_{(\mathbf{x}_{m},\mathbf{y}_{m})\in \mathcal{K}^{j}_{\mathbf{x}_{i}}}  \llbracket y_{ij}\neq y_{mj}  \rrbracket  &\quad y_{ij}=1 \\ 
0 & \text{otherwise}
\end{matrix}\right. 
\label{eq:lsd_localimbalance}
\end{equation}
where $\mathcal{K}^{j}_{\mathbf{x}_{i}}$ represents $k$NNs of $\mathbf{x}_{i}$ retrieved based on $l_j$-specific distance. 
A value of $\hat{b}_{ij}$ close to 0 (1) indicates a neighborhood of similarly (oppositely) labeled instances, denoting an internal (a boundary) area. In addition, $\hat{b}_{ij}$=1 implies that $\mathbf{x}_{i}$ is an outlier for $l_j$. 
As high-frequency (balanced) labels contain more ``1''s than low-frequent (imbalanced) ones, we apply column-wise normalization to $\hat{b}_{ij}$ to highlight the importance of instances relevant to highly imbalanced labels:
\begin{equation}
b'_{ij}= \frac{\hat{b}_{ij} \llbracket \hat{b}_{ij}\neq 1\rrbracket }{\sum_{i=1}^{n} \hat{b}_{ij} \llbracket \hat{b}_{ij}\neq 1\rrbracket}
\label{eq:lsd_localimbalance1}
\end{equation}
where $\llbracket \hat{b}_{ij}\neq 1\rrbracket$ indicates that outlier instances are excluded.
Finally, the local imbalance weight of $\mathbf{x}_{i}$ is computed by summing all  $b_{ij}$ of its relevant minority labels:
\begin{equation}
{u}^{b}_{i}=\sum_{l_j \in L_{\text{min}} } b'_{ij} y_{ij}\\
\label{eq:b_insweight}
\end{equation}
A higher weight indicates closer proximity to a more imbalanced label boundary.

The symmetric cosine similarity matrix $\mathbf{A}$ used in Section \ref{LSD} represents dependencies among label pairs. The higher value $A_{jh}$ is, the more correlated labels $l_j$ and $l_h$ are.
We obtain the strictly upper triangular matrix of $\mathbf{A}$ by setting all elements on and below the main diagonal as zero. Then, all non-zero elements are sorted in descending order. The label correlation matrix $\Tilde{\mathbf{A}}$ is obtained by retaining the top $tr\%$ elements and setting others to 0. Non-zero values $\Tilde{A}_{jh}$ indicate a strong correlation between labels $l_{j}$ and $l_{h}$, and we add $l_{j}$ and $l_h$ to ${L}_{h}^{cor}$ and ${L}_{j}^{cor}$ respectively, where ${L}_{j}^{cor}$ denotes the correlated label set of $l_j$ which captures label correlations in the global view, i.e., the whole instance set.
On the other hand, the local label distribution is more crucial to determine the importance of an instance \cite{MLSOL}. Therefore, we assess the label correlation carried by an instance for a specific label by considering both global and local label distributions. Specifically, an example $(\mathbf{x}_{i}, \mathbf{y}_{i}) \in {P}_{j}$ carries more label correlations, if more of its neighbors are relevant to both $l_j$ and $l_h$, where $l_h \in L_j^{cor}$. $L_j^{cor} = \emptyset$ indicates the nonexistence of significant global dependency between $l_j$ and other labels. Therefore, we define the label correlation information carried by $\mathbf{x}_{i}$ under its associated label $l_j$ as:
\begin{equation}
c_{ij} = \left\{
\begin{aligned}
& \mathop{\max}\limits_{{l}_{h} \in L_{j}^{cor}} \frac{y_{ih} + \sum_{\mathbf{x}_{m}\in \mathcal{K}^{j}_{\mathbf{x}_{i}}} y_{mh}}{k+1} , \quad \text{if }  L_{j}^{cor} \neq \emptyset \\
& 0, \quad \text{if }  L_{j}^{cor} = \emptyset
\end{aligned}
\right.
\label{eq:labelcorrelation}
\end{equation}
where labels in $L_j^{cor}$ share strong global label correlations with $l_j$, and the $max$ operator picks up the most locally correlated label of $l_j$.
The label correlation weight of $\mathbf{x}_{i}$ is computed by summing correlations carried by all relevant labels:
\begin{equation}
{u}^{c}_{i}=\sum_{l_j \in L_{min}}c_{ij}y_{ij}\\
\label{eq:c_insweight}
\end{equation}
The higher the weight, the more label correlation information an instance has. Finally, we derive the importance of the $i$-th sample in seed selection by combining its local imbalance and label correlation weights, i.e., $ u_{i}={u}^{b}_{i}+{u}^{c}_{i}$.

The process of instance weight calculation is illustrated in Algorithm \ref{al:Weight_Calculate}. 
We exemplify the effectiveness of instance weight used in LSDMLO with the toy dataset in Figure \ref{fig:boundary}.
The local imbalance weight of the internal example $\mathbf{x}_{3}$ is smaller than that of the two borderline examples $\mathbf{x}_{1}$ and $\mathbf{x}_{2}$ ($u_3^b<u_1^b$ and $u_3^b<u_2^b$). 
Additionally, $\mathbf{x}_{1}$, surrounded by instances relevant to both $l_1$ and $l_2$, carries more label correlation information and possesses higher label correlation weight than $\mathbf{x}_{2}$ and $\mathbf{x}_{3}$ ($u_2^c<u_1^c$ and $u_3^c<u_1^c$). For the outlier example $\mathbf{x}_{4}$ associated with $l_1$ only, its weight is 0. In summary, the order of instance weights is $u_{4}<u_{3}<u_{2}<u_{1}$.

\begin{algorithm}[t]
\caption{InsWeightCal}
\label{al:Weight_Calculate}
\SetKwData{Left}{left}\SetKwData{This}{this}\SetKwData{Up}{up}
\SetKwFunction{InitTypes}{InitTypes}\SetKwFunction{CreateIns}{CreateIns}
\SetKwInOut{Input}{input}\SetKwInOut{Output}{output}
\Input{multi-label training data set: $D$, label correlation matrix $\mathbf{A}$ , label-specific feature weight matrix $\mathbf{\Theta}$}
\Output{Instance weight vector $\mathbf{u}$}
\For{$l_{j}$ $\in$ $L_{min}$}
{
      Obtain ${P}_{j}=\left\{(\mathbf{x}_{i}, \mathbf{y}_{i})| y_{ij}=1, \ 1 \leq i \leq n \right\}$ \; 
      Obtain ${L}_{j}^{cor}$\ based on $\mathbf{A}$ \;    
      \For{$(\mathbf{x}_{i},\mathbf{y}_{i}) \in {P}_{j}$}
      {
      Get $\mathcal{K}^{j}_{\mathbf{x}_{i}}$ based on $l_j$-specific distance \; 
      Calculate local imbalance weight $\hat{b}_{ij}$ based on Eq.\eqref{eq:lsd_localimbalance}\;
      Calculate label correlation weight $c_{ij}$ based on Eq.\eqref{eq:labelcorrelation} \;
      }
}
Get $b'_{ij}$ by normalizing $\hat{b}_{ij}$ based on Eq.\eqref{eq:lsd_localimbalance1} for each instance and label \;
\For{$(\mathbf{x}_{i},\mathbf{y}_{i}) \in D$}
{
Calculate $u^b_i$ based on Eq.\eqref{eq:b_insweight} \;
Calculate $u^c_i$ based on Eq.\eqref{eq:c_insweight} \;
Calculate instance importance $u_i = u^b_i+u^c_i$ \;
}
\KwRet{$\mathbf{u} = [u_1, \cdots , u_n]$} 
\end{algorithm}

\subsection{Synthetic Instance Generation \label{IG}}

\begin{algorithm}[h]
\caption{InsGen}
\label{al:InstanceGeneration}
\SetKwData{Left}{left}\SetKwData{This}{this}\SetKwData{Up}{up}
\SetKwFunction{InitTypes}{InitTypes}\SetKwFunction{CreateIns}{CreateIns}
\SetKwInOut{Input}{input}\SetKwInOut{Output}{output}
\Input{multi-label dataset $D$, instance weight vector $\mathbf{u}$, minority label set $L_{min}$, sampling ratio: $p$}
\Output{new data set $D'$}
$D' \leftarrow D$ \;
\For{i $\leftarrow$ 1 to  $\lfloor np \rfloor$}
{
Seed instance $(\mathbf{x}_{s}, \mathbf{y}_{s}) \leftarrow $  roulette wheel selection based on $\mathbf{u}$ \;
Randomly choose a label $l_j$ from $\{l_h \in L_{min}|y_{sh}=1 \}$ \;
Randomly select a reference instance $(\mathbf{x}_{r}, \mathbf{y}_{r})$ from ${\mathcal{K}}^{j}_{\mathbf{x}_{s}}$\;
Get feature vector $\mathbf{x}_{g}$ based on Eq. \eqref{eq:sythetic_feature} \;
Get label vector $\mathbf{y}_{g}$ based on Eq. \eqref{eq:sythetic_label} \;
$D' \leftarrow D'\cup \{ (\mathbf{x}_{g},\mathbf{y}_{g}) \}$ \;
}
\KwRet{$D'$} 
\end{algorithm}

The procedure of synthetic instance generation is described in Algorithm \ref{al:InstanceGeneration}.
Given the sampling ratio $p$, the number of new instances to be generated is defined as $np$.
For each synthetic instance, a seed instance $(\mathbf{x}_{s}, \mathbf{y}_{s})$ is chosen based on instance weights $\mathbf{u}$ using the roulette wheel selection algorithm \cite{roulettealgorithm}, i.e., samples with a higher weight are more likely to be selected. 
Then, a minority label $l_j$ relevant to $\mathbf{x}_{s}$ is randomly chosen, and an arbitrary neighbor of ${\mathcal{K}}^{j}_{\mathbf{x}_{s}}$, denoted as $(\mathbf{x}_{r}, \mathbf{y}_{r})$, is selected as the reference instance. The randomness of the reference sample and relevant minority label selection guarantees the generation of diverse synthetic instances, avoiding the overfitting issue.


Let $\mathbf{x}_{g}\in\mathbb{R}^d$ be the feature vector of the synthetic instance generated upon the seed instance $(\mathbf{x}_{s}, \mathbf{y}_{s})$ and the reference instance $(\mathbf{x}_{r}, \mathbf{y}_{r})$. $x_{gi}$, which denotes the $i$-th feature of the synthetic instance, is calculated as: 
\begin{equation}
x_{gi}= \left\{
\begin{aligned}
& \alpha x_{si}+(1-\alpha) x_{ri} \  
\quad \text{numeric} \\
& \mathop{\arg\max}\limits_{v} \textstyle \sum_{\mathbf{x}_{h}\in (\mathbf{x}_{s} \cup {\mathcal{K}}^{j}_{\mathbf{x}_{s}} )}\llbracket x_{hi} =v  \rrbracket \    
\quad \text{nominal}
\end{aligned}
\right.
\label{eq:sythetic_feature}
\end{equation}
where $\alpha$ denotes a random number within the range of $[0,1],$ and $v$ represents all possible nominal values of the $i$-th feature. If $x_{gi}$ is a numerical feature, its value is a random point in the line that linearly interpolates the seed and reference instances. Otherwise, $x_{gi}$ is a nominal feature, and its value is set as the most common occurrence of the $i$-th feature value among $\mathbf{x}_{s}$ and its neighbors.

Concerning the label vector of synthetic instance $\mathbf{y}_{g}$, we propose a label-specific distance-induced label assignment strategy:
\begin{equation}
y_{gj}= \left\{
\begin{aligned}
& y_{sj}  \text{, if } y_{sj} = y_{rj} \\
& y_{ij}, \ i= \arg\min_{i \in \{s,r\}} d(\mathbf{x}_{g}, \mathbf{x}_{i}, \mathbf{\theta}_{\cdot j})  \text{, if } y_{sj} \neq y_{rj}
\end{aligned}
\right.
\label{eq:sythetic_label}
\end{equation}
Specifically, if a label $l_j$ is relevant (irrelevant) to both seed and reference instances, i.e., $y_{sj}=y_{rj}$, we directly assign $y_{sj}$ to $y_{gj}$. On the other hand, when $y_{sj}$ and $y_{rj}$ have different values, the label $l_j$ of the nearest (seed or reference) instance is assigned to the synthetic instance based on the $l_{j}$-specific distance.

LSDMLO generates more label-consistent and informative synthetic instances than existing neighborhood-based multi-label oversampling methods, as illustrated by the example in \textit{Appendix Figure A2}.

\subsection{Computational Complexity}
The time complexity of deriving label-specific distance weight (\textit{Appendix Algorithm 1}) is $O\left(t(nd^2 + d^2q + ndq + n^2d + dq^2) + d^3 +q^3 \right)$, where $t$ is the number of iterations (See \textit{Appendix 1.1} for details).  
The complexity of searching $k$NN using label-specific distance for all instances associated with minority labels is $O\left(nd\sum_{l_j \in L_{min}} n_j+|L_{min}|k^2\right)$, where $n_j$ denotes the number of instances associated with label $l_j$. 
The complexity of obtaining $\{L_j^{cor}\}_{j=1}^q$, $\mathbf{u}^b$ and $\mathbf{u}^c$ are $O(q^2\log q)$, $O\left(k\sum_{l_j \in L_{min}} n_j\right)$ and $O\left(k\sum_{l_j \in L_{min}} |L_j^{cor}|+n_j\right)$, respectively. 
Considering that $\sum_{l_j \in L_{min}} |L_j^{cor}|= \frac{1}{2}tr*q(q-1)$ and $k \ll n$, the complexity of computing the instance weight vector $\mathbf{u}$ (Algorithm \ref{al:Weight_Calculate}) is $O \left(nd\sum_{l_j \in L_{min}} n_j+|L_{min}|k^2 + q^2\log q \right)$.
The synthetic instance generation leads to a complexity of $O\left(pn(q+d)\right)$ (Algorithm \ref{al:InstanceGeneration}).
Due to $|L_{min}|<q$, $k \ll n$ and $np<n$, the overall complexity of LSDMLO is 
$O (t(nd^2 + d^2q + ndq + n^2d + dq^2)+  d^3 + q^3 + nd\sum_{l_j \in L_{min}} n_j )$.

\section{Experiments \label{sec:experiments} }
\subsection{Datasets and Experiment Setup \label{sec:experiment_setup}}
We conduct experiments on 13 multi-label datasets from various domains, and a comprehensive overview of their characteristics is presented in Table \ref{ta:Dataset}. 
All datasets utilized in our study are publicly available online  
\footnote{\url{http://mulan.sourceforge.net/}} \footnote{\url{http://www.uco.es/kdis/mllresources/}}  \footnote{\url{https://cometa.ujaen.es/}}.

\begin{table}[t]  
\caption{The characteristics of multi-label datasets, where Card (label cardinality) denotes the average number of labels per instance, and Dens (label density) is defined as Card divided by the total number of labels.} 
\centering
\renewcommand{\arraystretch}{1}
\resizebox{0.8\textwidth}{!}{
\begin{tabular}{@{}ccccccccccc@{}}
\toprule
Dataset & Domain & $n$ & $d$ & $q$ & Card & Dens & $MeanIR$ & $LImb$ \\ \midrule
emotions & music & 593 & 72 & 6 & 1.87 & 0.31 & 1.48 & 0.56  \\
scene & images & 2407 & 294 & 6 & 1.07 & 0.18 & 1.25 & 0.23 \\
yeast & biology & 2417 & 103 & 14 & 4.24 & 0.30 & 7.20 & 0.65  \\
Corel5k & images & 5000 & 499 & 374 & 3.52 & 0.01 & 189.57 & 0.97 \\
rcv1subset1 & text & 6000 & 472 & 101 & 2.88 & 0.03 & 54.49 & 0.90 \\
rcv1subset2 & text & 6000 & 472 & 101 & 2.63 & 0.03 & 45.51 & 0.89 \\
rcv1subset3 & text & 6000 & 472 & 101 & 2.61 & 0.03 & 68.33 & 0.89 \\
yahoo-Arts & text & 7484 & 231 & 25 & 1.67 & 0.07 & 26.00 & 0.86 \\
yahoo-Business & text & 11236 & 219 & 28 & 1.47 & 0.06 & 14.90 & 0.85 \\
cal500 & music & 502 & 68 & 174 & 26.04 & 0.15 & 19.38 & 0.84 \\
Corel16k & image & 13770 & 500 & 153 & 2.86 & 0.02 & 34.16 & 0.95 \\
Eurlex-sm & text & 19350 & 500 & 201 & 2.21 & 0.01 & 181.53 & 0.73 \\
tmc2007-500 & text & 28600 & 500 & 22 & 2.22 & 0.10 & 17.13 & 0.67 \\\bottomrule
\end{tabular}
}
\label{ta:Dataset}
\end{table}

We employ four widely used multi-label evaluation metrics, namely Macro-F, Macro-AUC, Macro-AUCPR, and Ranking Loss \cite{mllreview2}. The first three are label-based metrics that are sensitive to label imbalance and commonly used in imbalanced multi-label learning \cite{COCOA, MLSOL}. Ranking Loss is an example-based metric that measures the proportion of irrelevant labels ranked above relevant ones.
We conducted all experiments using $5\times 2$ cross-validation in combination with multi-label data stratification \cite{stratification}.
Our experiments were implemented based on Scikit-Multilearn library \cite{Scikit_MLL} and executed using Intel Xeon i5-119000 CPU and NVIDIA A5000 GPU. 
The source code of our method is available at \href{https://github.com/CquptZA/LSDMLO-PR-} {https://github.com/CquptZA/LSDMLO-PR-}.

We compare LSDMLO with nine state-of-the-art multi-label sampling methods, including two random sampling approaches (MLROS and MLRUS \cite{ML}), {RHwRSMT} \cite{REMEDIAL-HwR}, two variants of MLSMOTE (MLSMOTE\textsubscript{U} and MLSMOTE\textsubscript{R} that use union and ranking-based label assignment respectively \cite{MLSMOTE}),  MLSOL \cite{MLSOL}, MLBOTE \cite{MLSMOTE}, MLONC \cite{MLONC}, AEMLO \cite{AEMLO} and DR-SMOTE \cite{DR-SMOTE}. 
We consider four standard multi-label learners (BR \cite{BR}, MLkNN \cite{MLKNN}, RAkEL \cite{RAkEL},  CC \cite{ECC}), the imbalance-aware method COCOA \cite{COCOA}, and the deep learning model C2AE \cite{C2AE} as the base learners to combine with sampling approaches. 

The hyperparameters of the compared sampling methods and base multi-label learning models follow their original implementations and are detailed in \textit{Appendix Section A2.1}.
For LSDMLO, the number of neighbors $k=5$, label correlation retain ratio $tr=0.1$, and $\lambda _{1}$ and $\lambda _{2}$ are searched in $\{2^{-6},2^{-5}, \dots , 2^{6}\}$. Sampling ratio $p$ is chosen in $\{ 0.1, 0.3, \dots, 0.9\}$ for small datasets ($n \le 1000$), $\{ 0.1, 0.15, \dots, 0.3\}$ for medium datasets ($1000 < n \le  5000$), and $\{0.01,0.03, \cdots, \\ 0.09\}$ for large datasets ($n>5000$).

\subsection{Experiment Results  \label{result}}
Table \ref{ta:total} summarizes the average ranks of different sampling methods across various base classifiers, alongside Wilcoxon signed-rank test results with Bergman-Hommel correction \cite{Statisticaltesting} at the 5\% significance level. The detailed results are listed in \textit{Appendix Tables A1-A6}. The Base column denotes the results of the corresponding multi-label base classifier trained by the original (unsampled) datasets.

LSDMLO is the best sampling method for all base classifiers in terms of four evaluation metrics, achieving the highest average ranks. In addition, the proposed method obtains the most significant wins against compared approaches and does not suffer any significant loss. The success of LSDMLO is mainly due to the effectiveness of label-specific distance in retrieving class-consistent nearest neighbors and assigning proper labels.
MLSOL and MLBOTE are the top three methods in most cases. Although they emphasize borderline instances, the ignorance of label correlation and the reliance on Euclidean distance render them less effective than LSDMLO. 
AEMLO and MLONC, which explicitly exploit non-linear and pairwise label correlations comes next, but they do not incorporate label-specific discriminative features.
Other oversampling methods (MLROS, DR-SMOTE and MLSMOTE variants) improve base learners.
The undersampling method MLRUS is slightly worse than base classifiers, which implies that the removal of instances leads to inevitable information losses.
RHwRSMT performs worst in most cases, because decoupling one instance into two examples with distinct labels introduces extra noise in the decision boundaries.

\begin{table*}[h] 
\caption{The average rank and statistical tests under six base learners. The parenthesis ($n_1$/$n_2$) indicates the corresponding method is significantly superior to $n_1$ methods and inferior to $n_2$ methods based on the Wilcoxon signed rank test with Bergman-Hommel's correction at the 5\% level. The best methods are highlighted by \textbf{boldface}.}
\centering
\setlength{\tabcolsep}{2pt}
\resizebox{1.0\textwidth}{!}
{
\begin{tabular}{ccccccccccccccc}
\toprule
Metric & Base Learner & Base & MLRUS & RHwRSMT & MLROS & MLSMOTE\textsubscript{U} & MLSMOTE\textsubscript{R} & MLSOL & MLBOTE & MLONC & DR-SMOTE & AEMLO & LSDMLO \\ \midrule
\multirow{7}{*}{Macro-F} 
& BR    & 7.38(1/2)  & 8.31(1/2)  & 12.00(0/11) & 6.31(1/1)  & 9.15(1/4)  & 7.23(1/1)  & 4.38(2/0)  & 4.85(2/0)  & 4.62(2/1)  & 8.46(1/1)  & 3.54(3/0)  & \textbf{1.77(8/0)}  \\
& MLkNN & 8.62(1/3)  & 9.31(1/4)  & 11.85(0/11) & 6.15(2/0)  & 8.23(1/2)  & 5.77(1/0)  & 3.77(2/0)  & 4.54(3/0)  & 4.69(3/0)  & 5.92(1/0)  & 6.15(1/0)  & \textbf{3.00(4/0)}  \\
& CC    & 8.54(1/3)  & 8.85(1/3)  & 12.00(0/11) & 6.15(1/0)  & 9.00(1/3)  & 8.23(1/3)  & 4.92(1/0)  & 5.54(1/0)  & 2.92(5/0)  & 6.46(1/0)  & 2.77(5/0)  & \textbf{2.62(5/0)}  \\
& RAkEL & 9.23(1/6)  & 8.77(1/2)  & 12.00(0/11) & 5.08(2/0)  & 9.08(1/0)  & 8.00(1/1)  & 4.85(2/1)  & 4.77(2/0)  & 4.46(1/0)  & 6.46(2/0)  & 3.85(3/0)  & \textbf{1.46(5/0)}  \\
& COCOA & 8.46(0/2)  & 9.46(0/2)  & 11.23(0/7)  & 5.62(0/0)  & 8.77(1/1)  & 7.08(1/1)  & 4.77(2/0)  & 4.92(2/1)  & 6.46(0/0)  & 5.15(1/0)  & 4.54(1/1)  & \textbf{1.54(7/0)}  \\
& C2AE  & 9.77(0/3)  & 8.54(0/1)  & 10.54(0/3)  & 6.08(0/1)  & 8.15(0/0)  & 6.31(0/0)  & 5.69(0/0)  & 4.38(2/0)  & 6.08(0/1)  & 6.23(0/0)  & 4.31(2/1)  & \textbf{1.92(6/0)}  \\ \midrule
& \textit{Ave}(\textit{Total}) & 8.67(4/19) & 8.87(4/14) & 11.60(0/54) & 5.90(6/2) & 8.73(5/10) & 7.10(5/6) & 4.73(9/1) & 4.83(12/1) & 4.87(11/2) & 6.45(6/1) & 4.19(15/2) & \textbf{2.05(35/0)} \\ \hline

\multirow{7}{*}{\begin{tabular}[c]{@{}c@{}}Macro\\-AUC\end{tabular}}
& BR    & 8.38(0/3)  & 10.15(0/5) & 8.23(0/1)  & 7.92(0/2)  & 7.23(0/1)  & 7.62(0/3)  & 3.46(5/0)  & 3.62(3/1)  & 5.23(1/0)  & 9.92(0/4)  & 4.62(3/0)  & \textbf{1.62(8/0)}  \\
& MLkNN & 7.54(0/2)  & 10.15(0/3) & 9.54(0/2)  & 5.85(0/0)  & 8.38(0/2)  & 7.85(0/2)  & 3.69(0/0)  & 2.77(6/0)  & 4.69(1/0)  & 7.85(0/0)  & 7.31(0/2)  & \textbf{2.38(6/0)}  \\
& CC    & 7.92(1/1)  & 8.92(0/1)  & 11.46(0/9) & 6.54(1/1)  & 9.54(1/3)  & 6.69(0/1)  & 5.00(2/0)  & 4.46(2/1)  & 5.23(1/0)  & 6.38(1/0)  & 4.23(1/0)  & \textbf{1.62(7/0)}  \\
& RAkEL & 9.77(0/5)  & 10.38(0/4) & 9.46(0/1)  & 6.31(1/0)  & 8.15(0/1)  & 6.69(1/1)  & 4.08(0/0)  & 4.92(1/1)  & 3.85(2/0)  & 8.15(0/2)  & 4.15(3/0)  & \textbf{2.08(7/0)}  \\
& COCOA & 7.23(1/0)  & 10.15(0/6) & 11.54(0/8) & 6.08(1/1)  & 8.62(0/1)  & 6.46(1/1)  & 4.46(2/0)  & 4.77(2/0)  & 6.85(1/0)  & 5.69(2/1)  & 4.08(2/0)  & \textbf{2.08(6/0)}  \\
& C2AE  & 10.15(0/5) & 9.46(0/3)  & 10.15(0/4) & 5.38(1/0)  & 8.00(0/1)  & 8.62(0/4)  & 4.08(3/0)  & 4.23(4/0)  & 4.38(2/0)  & 7.31(0/1)  & 4.46(2/1)  & \textbf{1.77(7/0)}  \\ \midrule
& \textit{Ave}(\textit{Total}) & 8.50(2/16) & 9.87(0/22) & 10.06(0/25) & 6.35(4/4) & 8.32(1/9) & 7.32(2/12) & 4.13(12/0) & 4.13(18/3) & 5.04(8/0) & 7.55(3/8) & 4.81(11/3) & \textbf{1.92(41/0)} \\ \hline

\multirow{7}{*}{\begin{tabular}[c]{@{}c@{}}Macro\\-AUPR\end{tabular}}
& BR    & 7.92(0/2)  & 8.15(0/2)  & 10.54(0/2)  & 6.08(0/0)  & 9.92(0/3)  & 7.23(0/1)  & 4.92(1/1)  & 3.92(2/0)  & 5.38(0/0)  & 8.38(0/2)  & 3.85(3/0)  & \textbf{1.69(7/0)}  \\
& MLkNN & 8.62(0/2)  & 9.77(0/3)  & 11.08(0/5) & 6.77(0/0)  & 9.15(0/1)  & 6.92(0/0)  & 4.62(0/0)  & 4.54(2/0)  & 4.69(1/0)  & 4.77(1/0)  & 4.69(3/0)  & \textbf{2.38(4/0)}  \\
& CC    & 7.38(0/1)  & 8.46(0/3)  & 10.85(0/5) & 5.62(3/0)  & 10.54(0/5) & 7.69(0/1)  & 3.54(2/0)  & 4.38(3/0)  & 6.00(0/0)  & 8.23(0/2)  & 3.54(3/0)  & \textbf{1.77(6/0)}  \\
& RAkEL & 8.15(0/0)  & 9.15(0/3)  & 11.00(0/6) & 6.77(0/0)  & 9.31(0/2)  & 7.46(1/2)  & 4.77(2/0)  & 4.00(3/0)  & 5.62(0/0)  & 6.54(1/1)  & 3.38(2/0)  & \textbf{1.85(5/0)}  \\
& COCOA & 6.92(0/1)  & 8.00(0/1)  & 11.00(0/6) & 5.92(1/0)  & 10.46(0/4) & 8.08(1/3)  & 4.31(3/0)  & 4.54(3/0)  & 5.92(0/0)  & 5.54(1/0)  & 5.46(1/0)  & \textbf{1.85(5/0)}  \\
& C2AE  & 9.85(0/5)  & 8.23(0/2)  & 10.00(0/4) & 6.38(0/1)  & 9.46(0/4)  & 6.46(0/0)  & 3.62(3/0)  & 3.85(4/0)  & 5.85(1/0)  & 7.54(0/1)  & 4.92(3/0)  & \textbf{1.85(6/0)}  \\ \midrule
& \textit{Ave}(\textit{Total}) & 8.14(0/11) & 8.63(0/14) & 10.74(0/28) & 6.26(4/1) & 9.81(0/19) & 7.31(2/7) & 4.29(11/1) & 4.21(17/0) & 5.58(2/0) & 6.83(3/6) & 4.31(15/0) & \textbf{1.90(33/0)} \\ \hline

\multirow{7}{*}{\begin{tabular}[c]{@{}c@{}}Ranking\\Loss\end{tabular}}
& BR    & 8.08(0/2)  & 8.77(0/1)  & 9.15(0/1)  & 6.46(0/0)  & 6.77(0/1)  & 8.23(0/2)  & 4.54(0/0)  & 4.62(1/0)  & 7.08(0/0)  & 8.00(0/0)  & 3.77(1/0)  & \textbf{2.54(5/0)}  \\
& MLkNN & 6.69(1/0)  & 6.69(0/0)  & 10.54(0/5) & 4.77(1/0)  & 8.31(0/0)  & 8.85(0/0)  & 6.00(0/0)  & 6.77(1/0)  & 5.23(0/0)  & 7.54(0/1)  & 4.15(1/0)  & \textbf{2.46(2/0)}  \\
& CC    & 6.23(0/0)  & 8.77(0/0)  & 10.92(0/4) & 6.54(1/0)  & 9.31(0/3)  & 7.77(0/0)  & 4.69(0/0)  & 4.15(2/0)  & 6.77(0/0)  & 7.15(0/2)  & 2.92(3/0)  & \textbf{2.77(3/0)}  \\
& RAkEL & 7.77(0/1)  & 8.92(0/2)  & 10.46(0/2) & 7.31(0/0)  & 8.15(0/1)  & 8.69(0/1)  & 3.85(0/0)  & 4.23(0/0)  & 5.00(0/0)  & 7.15(0/1)  & 4.31(2/0)  & \textbf{2.15(6/0)}  \\
& COCOA & 9.31(0/1)  & 9.23(0/1)  & 8.00(0/0)  & 5.31(0/0)  & 7.38(0/0)  & 8.62(0/2)  & 4.92(0/0)  & 4.08(1/0)  & 6.62(0/1)  & 8.08(0/2)  & 4.31(2/0)  & \textbf{2.15(4/0)}  \\
& C2AE  & 9.15(0/3)  & 7.69(0/1)  & 9.54(0/1)  & 6.38(0/0)  & 7.85(0/1)  & 8.54(0/3)  & 4.46(2/1)  & 3.85(2/0)  & 6.46(0/1)  & 7.92(0/1)  & 4.62(0/0)  & \textbf{1.54(8/0)}  \\ \midrule
& \textit{Ave}(\textit{Total}) & 7.87(1/7) & 8.35(0/5) & 9.77(0/13) & 6.13(2/0) & 7.96(0/6) & 8.45(0/8) & 4.74(2/1) & 4.62(7/0) & 6.19(0/2) & 7.64(0/7) & 4.01(9/0) & \textbf{2.27(28/0)} \\ \bottomrule
\end{tabular}
}
\label{ta:total}
\end{table*}

The effectiveness of sampling methods varies in terms of different metrics.
In Table \ref{ta:total}, sampling methods significantly improve the six base learners in 19 and 16 cases (base learners sustain 19 and 16 significant losses) in terms of Macro-F and Macro-AUC respectively. On the other hand, the six base learners obtain only 11 significant losses against all sampling methods in terms of Macro-AUPR. This observation is consistent with \cite{MLSOL}, as Macro-AUPR is more sensitive to the precision of high-ranked minority class instances, which makes it more challenging to be enhanced than the other two Macro-averaged metrics, especially in extremely imbalanced cases. 
Nevertheless, LSDMLO still significantly elevates base learners in terms of all three Macro-averaged metrics.
Regarding Ranking Loss, the six base learners are significantly inferior to sampling methods in only 7 cases, and MLkNN and CC do not even suffer any significant loss. Ranking Loss is an example-based metric that assesses the portion of relevant labels ranking lower than irrelevant labels. As the number of majority labels is typically larger than that of minority labels in each instance, the majority label is more important in Ranking Loss. On the other hand, the macro-averaged label metrics aggregate the performance of every label, which are dominated by minority labels. All sampling methods aim to improve the performance of minority (imbalanced) labels, thereby being more effective in label-based metrics than Ranking Loss. Nevertheless, LSDMLO exploits correlation among labels to encourage correlated relevant labels to rank higher simultaneously, leading to its superiority over other competitors in terms of Ranking Loss.


The results of the sampling methods depend on the base multi-label learners they are paired with. The better the base learner, the higher the performance of the sampling methods. 
C2AE and COCOA are the top two base learners, largely exceeding other traditional base learners in terms of the four metrics. 
Therefore, we further investigate the performance of the sampling methods when combined with C2AE and COCOA.
C2AE, a deep multi-label learning model, is the best-performing learner. LSDMLO achieves 26/52 top results in terms of four metrics, followed by MLSOL which ranks first in 6 cases. COCOA, an imbalance-aware multi-label classifier, is the second-best base learner. LSDMLO is the best sampling method in 25/52 cases, and MLSOL which obtains 8 best results comes next. Therefore, the proposed LSDMLO is more effective in improving better-performed multi-label base learners. 

\subsection{Effectiveness of Label-Specific Distance \label{sectionlimb}}

$LImb$ measures label imbalance within local regions rather than in the whole instance space, which is a useful metric to assess the proportion of neighbors associated with the same labels and evaluate the difficulty of multi-label datasets \cite{MLSOL}. 
Specifically, a smaller $LImb$ implies that more neighbors share the same labels and the dataset is easier to learn. 
Apart from $LImb$ defined upon Euclidean distance-based neighbors (Eq.\eqref{LImb}), we further compute local imbalance based on label-specific distances derived neighbors:
\begin{equation}
LImb_{LSD}=\frac{1}{q}\displaystyle\sum_{j=1}^{q}\frac{\sum_{i=1}^{n}\hat{b}_{ij}y_{ij}}{\sum_{i=1}^{n}y_{ij}}
\label{eq:LImb_LSD}
\end{equation}

\begin{figure*}[!h]
\centering
\includegraphics[width=0.75\textwidth]{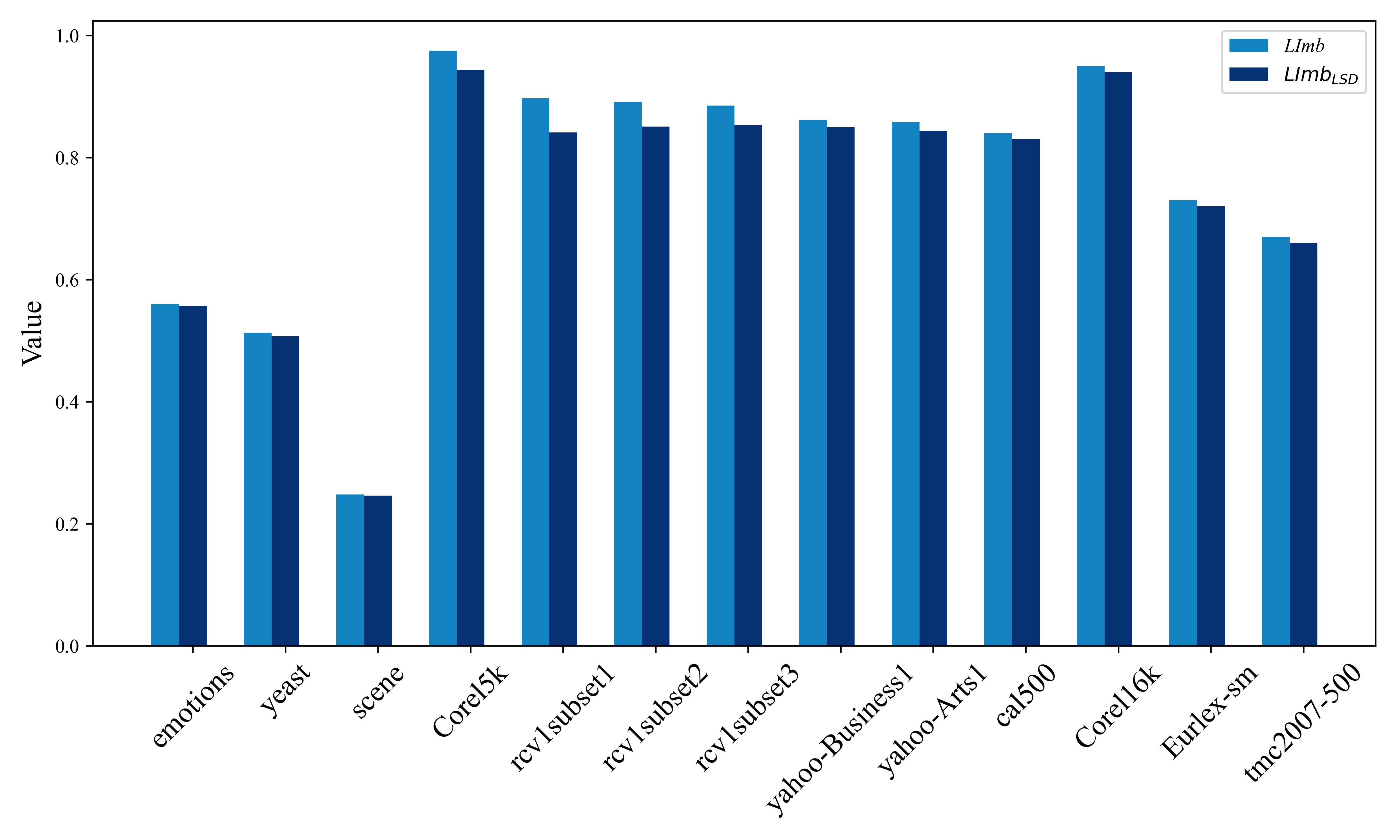} 
\caption{The comparison of local label imbalance based on Euclidean distance and label-specific distance. }
\label{fig:LImb}
\end{figure*}

As shown in Figure \ref{fig:LImb}, $LImb_{LSD}$ is smaller than $LImb$ on all datasets, demonstrating the effectiveness of label-specific distance to identify more label-consistent neighbors. Particularly, the decrease of $LImb_{LSD}$ is more significant for harder datasets ($LImb > 0.8$), which supports the utility of label-specific distance for complex local label distributions. Overall, label-specific distance is more powerful than Euclidean distance in identifying label-consistent neighborhoods under bewildering label distributions. 

Additional empirical studies, including comparisons with LLSF, parameter analysis, and computational evaluation, are presented in \textit{Appendix Sections A2.2, A2.3, and A2.4}, respectively.

\section{Conclusion}
We proposed LSDMLO, a multi-label oversampling method that derives label-specific distances to identify label-consistent neighbors via semantic-aware feature weighting. The label-specific distance-based neighborhood enables LSDMLO to select borderline seed instances possessing abundant label correlations and assign proper labels to synthetic instances.
The experimental results from benchmark datasets demonstrate the significant advantages of the proposed method over competitors, especially when collaborating with better-performing base classifiers. 

Future work will focus on developing a more comprehensive label-specific distance learning framework by incorporating global feature importance \cite{LDL} and instance correlations \cite{CLML}, and nonlinear modeling to better capture complex feature-label interactions beyond the current sparse linear formulation. 
We also plan to explore advanced label dependency modeling strategies capable of capturing higher-order label relationships beyond cosine similarity-based pairwise correlations.



\begin{credits}
\subsubsection{\ackname} This work was supported by the National Natural Science Foundation of China (62302074), Natural Science Foundation of Chongqing (CSTB2025NSCQ-GPX1269), the Science and Technology Research Program of Chongqing Municipal Education Commission (KJQN202300631).
\end{credits}


%
%
%
\bibliographystyle{splncs04}
\bibliography{ECML2026-CameraReady/ref}
%





\end{document}


\title{Addressing Imbalance in Multi-Label Data via Label-Specific Distance-based Oversampling --Appendix}

\titlerunning{MLDSLO Appendix}

\author{Bin Liu\inst{1} \and
Jun Wu\inst{1,2} \and
Haoyu Peng\inst{1,2} \and
Ao Zhou\inst{3} \corr \and
Jin Wang\inst{1} \and
QiaoSong Chen\inst{1} \and
Grigorios Tsoumakas\inst{4}
}


\authorrunning{B. Liu et al.}


\institute{Key Laboratory of Data Engineering and Visual Computing, Chongqing University of Posts and Telecommunications, China \email{\{liubin,wangjin,chenqs\}@cqupt.edu.cn}
\and
School of Computer Science and Technology, Chongqing University of Posts and Telecommunications, China
\email{\{s240231056,s240231185\}@stu.cqupt.edu.cn}
\and
State Key Laboratory of Novel Software Technology, Nanjing University, China \email{zacqupt@gmail.com}
\and
School of Informatics, Aristotle University of Thessaloniki, Greece 
\email{greg@csd.auth.gr}
}

\maketitle  

\appendix
\renewcommand{\thesection}{A\arabic{section}}
\renewcommand{\thesubsection}{A\arabic{section}.\arabic{subsection}}

\renewcommand{\thetable}{A\arabic{table}}  
\renewcommand{\thefigure}{A\arabic{figure}}  
\setcounter{table}{0}  
\setcounter{figure}{0} 

\section{Supplementary Description of LSDMLO}\label{app_sec_method}


\subsection{Optimization of Label-Specific Weights}

The label-specific feature weights matrix is obtained by solving the following objective:
\begin{equation}
\label{eq:lsd_obj_appendix}
\mathop{min}\limits_{\mathbf{W}}\frac{1}{2}\left \| \mathbf{XW}-\mathbf{Y}\right \|_{F}^{2}+\lambda _{1} \left \| \mathbf{W}\right \|_{1} +\frac{\lambda _{2}}{2}tr(\mathbf{XW}\mathbf{\Lambda}(\mathbf{XW})^{\top})
\end{equation}
Although Eq.\eqref{eq:lsd_obj_appendix} is inherently convex, introducing the \textbf{$\ell_{1}$} norm-regularizer renders it a non-smooth optimization challenge. 
To effectively address this non-smoothness and tackle the \textbf{$\ell_{1}$} norm-regularization, We embraces the accelerated proximal gradient (APG) method \cite{APG}. 
Within the APG framework, the convex optimization problem Eq.\eqref{eq:lsd_obj_appendix} is reformulated as:

\begin{equation}
\Theta(\mathbf{W})=\mathop{min}\limits_{\mathbf{W}\in \mathcal H} \Phi_{1}  +\Phi_{2} \quad 
\end{equation}
\begin{equation}
\Phi_{1} =\frac{1}{2}\left \| \mathbf{XW}-\mathbf{Y}\right \|_{F}^{2}+\frac{\lambda _{2}}{2}tr(\mathbf{XW}\mathbf{\Lambda}(\mathbf{XW})^{\top})
\label{Phi1}
\end{equation}
\begin{equation}
\Phi_{2}=\lambda _{1}\left \|\mathbf{W}\right \|_{1} 
\end{equation}
where $\mathcal H$ is a real Hilbert space. 
$\Phi_{1}$ is the smooth part of $\Theta$ and satisfies the lipschitz continuity condition, i.e., $(\left \| \triangledown \Phi_{1}(\mathbf{W}_{1}) - \triangledown \Phi_{1}(\mathbf{W}_{2}) \right \|_{F}^{2} \leq L_{\Phi_{1}} \left \| \mathbf{W}_{1}- \mathbf{W}_{2} \right \| )$ \footnote{$L_{\Phi_{1}}$ is the lipschitz constant}. The second term $\Phi_{2}$ enforces sparsity on $\mathbf{W}$ using \textbf{$\ell_{1}$} norm regularization with a coefficient $\lambda_{1}$.  Rather than minimizing $\Theta(\mathbf{W})$ directly, proximal gradient algorithms minimize a series of separable quadratic approximations to $\Theta(\mathbf{W})$, referred to as $Q(\mathbf{W}, \mathbf{W}^{(t)})$:
\begin{equation}
\begin{split}
Q(\mathbf{W}, \mathbf{W}^{(t)}) &= \Phi_{1}(\mathbf{W}^{(t)}) + \nabla \Phi_{1}(\mathbf{W}^{(t)})(\mathbf{W} - \mathbf{W}^{(t)}) \\
&\quad + L_{\Phi_{1}} \lVert \mathbf{W} - \mathbf{W}^{(t)} \rVert_{F}^{2} + \Phi_{2}(\mathbf{W})
\end{split}
\end{equation}
Let $\mathbf{G}^{(t)}= \mathbf{W}^{(t)}-\frac{1}{L_{\Phi_{1}}} \triangledown\Phi_{1}(\mathbf{W}^{(t)})$, where $\mathbf{W}^{(t)}$ is the result of $\mathbf{W}$ at the $t$-th iteration. $\mathbf{W}$ can be optimized by:

\begin{equation}
\begin{split}
\mathbf{W^*} &= \arg \min_\mathbf{W} Q(\mathbf{W}, \mathbf{W}^{(t)}) \\
&= \arg \min_\mathbf{W} \frac{L_{\Phi_{1}}}{2} \left\| \mathbf{W} - \mathbf{G}^{(t)} \right\|_{F}^{2} + \Phi_{2}(\mathbf{W}) \\
&= \arg \min_\mathbf{W} \frac{1}{2} \left\| \mathbf{W} - \mathbf{G}^{(t)} \right\|_{F}^{2}  + \frac{\lambda_{2}}{L_{\Phi_{1}}} \left\| \mathbf{W} \right\|_1
\end{split}
\end{equation}

The convergence rate can reduce to $O(t^{-2})$ by employing the setting $\mathbf{W}^{(t)} = \mathbf{W}_{t} + \frac{b_{t-1}-1}{b_{t}}(\mathbf{W}_{t}-\mathbf{W}_{t-1})$, where $b_{t}$ satisfies $b_{t+1}^{2} - b_{t+1} < b_{t}^{2}$. 
In each iteration, we address it by solving the following optimization problem.
\begin{equation}
\mathbf{W}_{t+1}=S_{\varepsilon }\left [ \mathbf{G}^{(t)}\right ]=\mathop{\arg \min}\limits_{\mathbf{W}}\varepsilon \left \| \mathbf{W}\right \|_{1}+\frac{1}{2}\left \| \mathbf{W}-\mathbf{G}^{(t)}\right \|_{F}^{2}
\end{equation}
where $S_{\varepsilon}$ is the soft-thresholding operator. For each $w_{ij}$ and $\varepsilon$ $>$ 0, the soft-thresholding operation is defined as:
\begin{equation}
S_{\varepsilon}[\omega]=\left\{\begin{matrix}
\omega - \varepsilon & \quad \text{if} \quad \omega>\varepsilon\\ 
\omega -\varepsilon  & \quad \text{if} \quad \omega<-\varepsilon\\ 
 0& \quad \text{otherwise}
\end{matrix}\right.  
\end{equation}
The accelerated proximal gradient algorithm to obtain optimal $\mathbf{W}$ is summarized in Appendix Algorithm \ref{al:W}.

\begin{algorithm}
\SetAlgorithmName{Appendix Algorithm}{Appendix Algorithm} 

\SetKwData{Left}{left}\SetKwData{This}{this}\SetKwData{Up}{up}
\SetKwFunction{InitTypes}{InitTypes}\SetKwFunction{CreateIns}{CreateIns}
\SetKwInOut{Input}{input}\SetKwInOut{Output}{output}
\Input{feature matrix: $\mathbf{X}$, label matrix: $\mathbf{Y}$, trade-off parameters $\lambda_{1}$ and $\lambda_{2}$ }
\Output{coefficient matrix $\mathbf{W}$}
$b_{0},b_{1} \leftarrow 1$;  $t \leftarrow 1$; $\gamma$=0.01\; 
$\mathbf{W}_{0}, \mathbf{W}_{1}\leftarrow 
(\mathbf{X}^{T}\mathbf{X}+\gamma \mathbf{I})^{-1}\mathbf{X}^T\mathbf{Y}$ 
\tcc*[r]{Initialization}
Calculate Lipschitz constant $L_{\Phi_{1}}$ by Eq.\eqref{Lipschitz}\;
\Repeat{convergence}
{
    $\mathbf{W}^{(t)}\leftarrow \mathbf{W}_{t}+\frac{b_{t-1}-1}{b_{t}}(\mathbf{W}_{t}-\mathbf{W}_{t-1})$\;
    $\mathbf{G}^{(t)}\leftarrow \mathbf{W}^{(t)}-\frac{1}{L_{\Phi_{1}}} \triangledown \mathbf{\phi_{1}}(\mathbf{W}^{(t)})$\;
    $\mathbf{W}_{t+1}\leftarrow S_{\frac{\lambda_{2}}{L_{\Phi_{1}}}}(\mathbf{G}^{(t)})$\;
    $b_{t+1}\leftarrow \frac{1+\sqrt{4b_{t}^{2}+1}}{2}$\;
    $ t \leftarrow t+1$ \;
}
$\mathbf{W} \leftarrow \mathbf{W}_{t+1}$  \;  
\KwRet{$\mathbf{W}$}\;
\caption{FeatureWOpti}
\label{al:W}
\end{algorithm}

\textbf{Proof of Lipschitz Continuity.} 
To prove that $L_{\Phi_{1}}$ is a Lipschitz constant means to demonstrate that the gradient of the function $\Phi_{1}$ satisfies the following Lipschitz condition for any $\mathbf{W}_{1}$ and $\mathbf{W}_{2}$:
\begin{equation}
\left\| \nabla \Phi_{1}(\mathbf{W}_{1}) - \nabla \Phi_{1}(\mathbf{W}_{2}) \right\|_{F}^2 \leq L_{\Phi_{1}} \left\| \mathbf{W}_{1} - \mathbf{W}_{2} \right\|_{F}  
\end{equation}
Given $\Phi_{1}$ as Eq \ref{Phi1},its gradient $\nabla \Phi_{1}(\mathbf{W})$ can be calculated as:
\begin{equation}
\begin{split}
 &\nabla \Phi_{1}(\mathbf{W}) = \\ 
 &\mathbf{X}^{\top}(\mathbf{XW} - \mathbf{Y}) + \lambda_{2} \mathbf{X}^{\top}\mathbf{XW} \mathbf{\Lambda}     
\end{split}
\end{equation}
Therefore, for any $\mathbf{W}_{1}$ and $\mathbf{W}_{2}$, the Frobenius norm of the gradient difference can be expressed as:
\begin{equation}
\begin{split}
&\left\| \nabla \Phi_{1}(\mathbf{W}_{1}) - \nabla \Phi_{1}(\mathbf{W}_{2}) \right\|_{F}^{2} = \\
&
\left\| \mathbf{X}^{\top}\mathbf{X}(\mathbf{W}_{1} - \mathbf{W}_{2}) \right\|_{F}^{2} + \lambda_{2} \left\| \mathbf{X}^{\top}\mathbf{X}(\mathbf{W}_{1} - \mathbf{W}_{2}) \mathbf{\Lambda} \right\|_{F}^{2}
\end{split}
\end{equation}
Utilizing properties of matrix norms and the maximum singular value $\sigma_{\max}$, we obtain:
\begin{equation}
\begin{split}
&\left\| \nabla \Phi_{1}(\mathbf{W}_{1}) - \nabla \Phi_{1}(\mathbf{W}_{2}) \right\|_{F}^2 \leq \\
&(\sigma_{\max}^{2}(\mathbf{X}^{\top}\mathbf{X}) + \lambda_{2} \sigma_{\max}^{2}(\mathbf{\Lambda}) \left\| \mathbf{W}_{1} - \mathbf{W}_{2} \right\|_{F}^{2}  
\end{split}
\end{equation}
indicating that the Lipschitz constant $L_{\Phi_{1}}$ can be defined as:
\begin{equation}
L_{\Phi_{1}} = \sqrt{\sigma_{\max}^{2}(\mathbf{X}^{\top}\mathbf{X}) +  \sigma_{\max}^{2}\lambda_{2}(\mathbf{\Lambda})}  
\label{Lipschitz}
\end{equation}


\textbf{Computational Complexity.} 
The time complexity of initializing $\mathbf{W}_0$ and $\mathbf{W}_1$ is $O(nd^2 + d^3 + ndq + d^2q)$.
The time complexity for calculating label correlation using cosine similarity is $O(nq^2)$. The initialization of Lipschitz constant $L_{\Phi_{1}}$ has a time complexity of $O(d^3 + q^3)$. 
In the iteration step, the time complexity of calculating the gradient $\phi_{1}$ is $O(nd^2 + d^2q + ndq + n^2d + dq^2)$. 
Therefore, the computational complexity of Appendix Algorithm \ref{al:W} can be represented as $O\left(t(nd^2 + d^2q + ndq + n^2d + dq^2) + d^3 +q^3 \right)$, where $t$ is the number of iterations. 

\subsection{Demonstrating the Advantage of LSDMLO}

\begin{figure*}[h]
\centering
\begin{subfigure}{0.3\textwidth}
    \includegraphics[width=\textwidth]{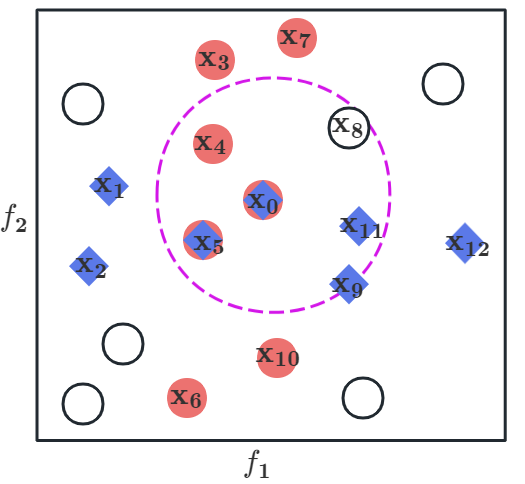}
    \caption{Euclidean distance}
\end{subfigure}
\begin{subfigure}{0.3\textwidth}
\includegraphics[width=\textwidth]{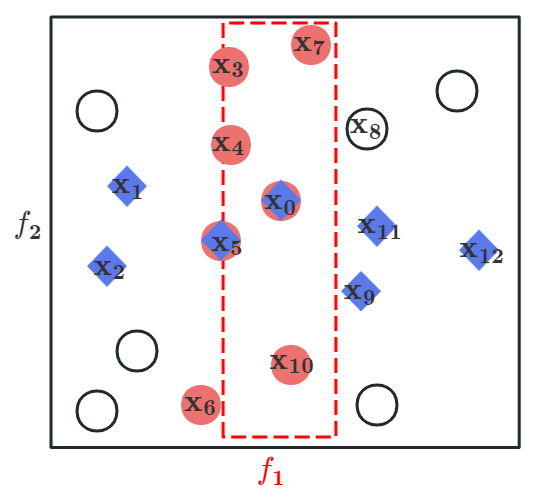}
\caption{$l_1$-specific distance}
\end{subfigure}
\begin{subfigure}{0.3\textwidth}
\includegraphics[width=\textwidth]{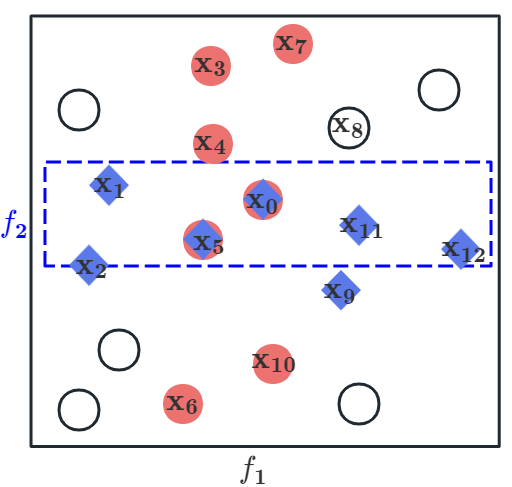}
\caption{$l_2$-specific distance}
\end{subfigure}
\begin{minipage}{0.75\linewidth}
\centerline{\includegraphics[width=1\textwidth]{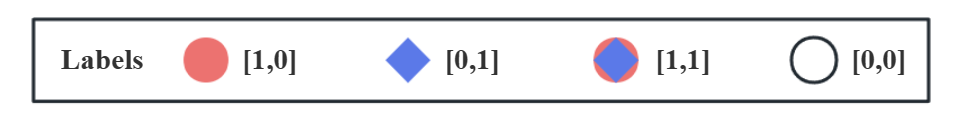}}
\end{minipage}
\caption{5NNs of $\mathbf{x}_0$ retrieved using (a) Euclidean distance, (b) $l_1$-specific distance ($\bm{\theta}_{\cdot 1}=[1,0]^\top$), and (c) $l_2$-specific distance ($\bm{\theta}_{\cdot 2}=[0,1]^\top$), where features $f_1$ and $f_2$ are more pertinent to $l_1$ and $l_2$, respectively.}
\label{fig:lsd}
\end{figure*}

Appendix Figure \ref{fig:lsd} shows an example of retrieving neighbor instances based on different distances. Only 2 out of 5 neighbors of $\mathbf{x}_0$ (in the purple circle) are associated with $l_1$ in Figure (a), while all neighbors defined by the $l_1$-specific distance (in the red rectangle) are relevant to $l_1$ in Figure (b). Similarly, $l_2$-specific distance also identifies five neighbors associated with $l_2$ (in the blue rectangle) in Figure (c). Compared to Euclidean distance, label-specific distance demonstrates enhanced discriminative capability in identifying neighbors assigned to the same label.

We illustrate the advantage of label-specific distance-aware synthetic instance generation used in LSDMLO via the toy dataset in Figure \ref{fig:lsd}. Let $\mathbf{x}_0$ be the seed instance, and its neighbors defined by Euclidean and $l_1$-specific distances are extracted and zoomed in Appendix Figure \ref{fig:labelassign}. 
MLSMOTE, MLBOTE and MLSOL leverage Euclidean distance-based neighbors ($\mathcal{K}_{\mathbf{x}_0}=\{\mathbf{x}_4, \mathbf{x}_5, \mathbf{x}_8, \mathbf{x}_{9}, \mathbf{x}_{11}\}$) to create new synthetic instances. 
MLSMOTE with the ranking strategy assigns a label to a synthetic instance if more than half neighbors are relevant to that label. Thus, all synthetic examples in MLSMOTE are associated with $l_2$, but none is relevant to $l_1$.
For MLBOTE, $\mathbf{x}_5$ that has the same label vector as $\mathbf{x}_s$ is the sole reference example, so all synthetic instances are confined in the line connecting $\mathbf{x}_s$ and $\mathbf{x}_5$, as shown in MLBOTE, leading to instance homogeneity and overfitting.
MLSOL determines labels of synthetic instances based on their locations, but it may suffer the risk of invalid example generation. As shown in Figure MLSOL, the label vector of $\mathbf{s}_2$ is the same as its closest instance $\mathbf{x}_8$ ($[0,0]$), but the creation of $\mathbf{s}_2$ fails to alleviate the imbalance for both $l_1$ and $l_2$.

\begin{figure*}[h]
\captionsetup[subfigure]{justification=centering, font=small}
\setlength{\tabcolsep}{2pt}

\begin{subfigure}{0.3\textwidth}
\includegraphics[width=\textwidth]{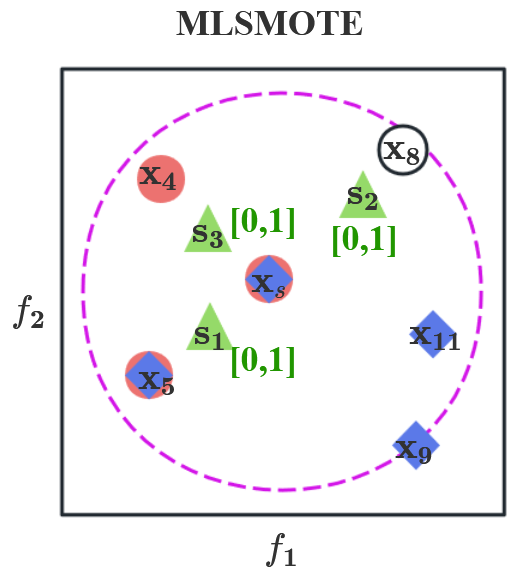}
\caption{MLSMOTE}
\end{subfigure}
\hfill 
\begin{subfigure}{0.3\textwidth}
\includegraphics[width=\textwidth]{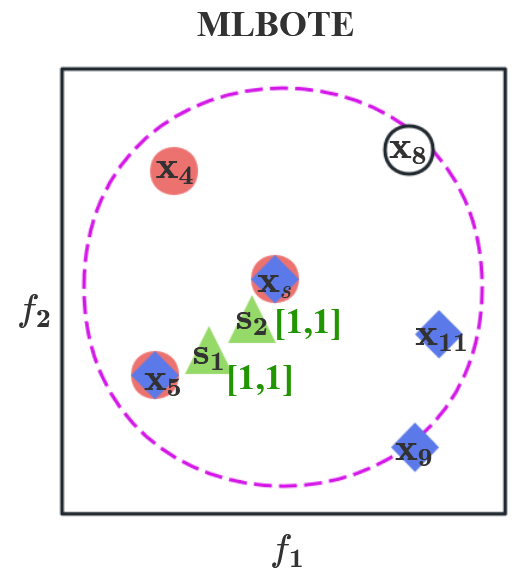}
\caption{MLBOTE}
\end{subfigure}
\hfill
\begin{subfigure}{0.3\textwidth}
\includegraphics[width=\textwidth]{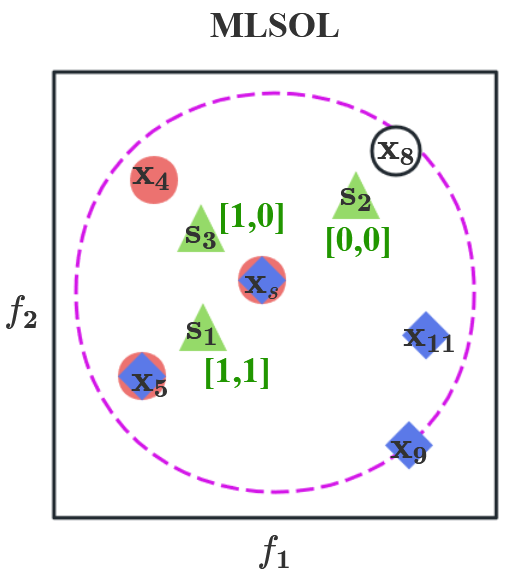}
\caption{MLSOL}
\end{subfigure}
\begin{subfigure}{0.3\textwidth}
\includegraphics[width=\textwidth]{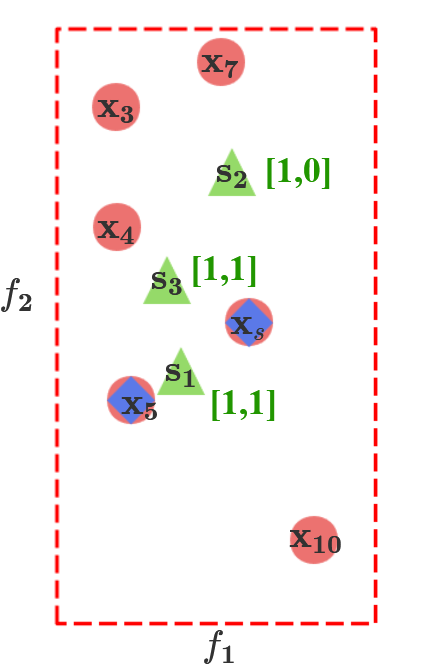}
\caption{LSDMLO}
\end{subfigure}
\hfill
\begin{subfigure}{0.3\textwidth} %
\includegraphics[width=\textwidth]{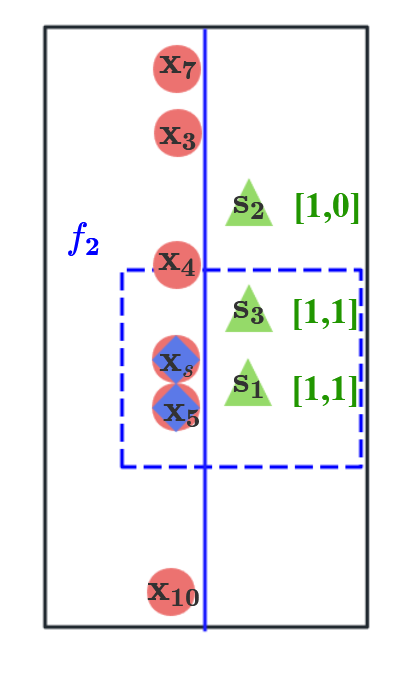}
\caption{$l_2$-specific distance}
\end{subfigure}
\hfill
\begin{subfigure}{0.3\textwidth} %
\centering 
\includegraphics[width=0.55\textwidth]{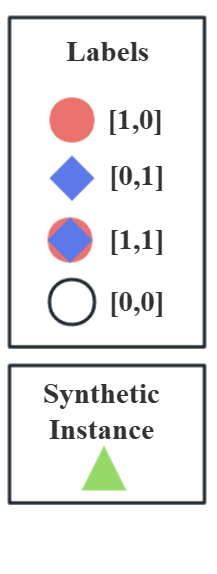} 
\caption{Legend}
\end{subfigure}

\caption{Comparison of synthetic instances created by LSDMLO and other oversampling methods.}
\label{fig:labelassign}
\end{figure*}

On the other hand, the proposed LSDMLO can tackle the limitations of the previous oversampling approaches. As shown in LSDMLO, LSDMLO generates $\mathbf{s}_1$ and $\mathbf{s}_3$ associated with both $l_1$ and $l_2$, influencing more labels than MLSMOTE and exhibiting more generalization than MLBOTE. 
Compared to MLSOL, LSDMLO avoids creating useless synthetic examples due to the high quality of the reference candidate set $\mathcal{K}^1_{\mathbf{x}_s}=\{\mathbf{x}_3, \mathbf{x}_4, \mathbf{x}_5, \mathbf{x}_{7}, \mathbf{x}_{10}\}$. 
With regard to the synthetic instance $\mathbf{s}_3$ created based on seed $\mathbf{x}_s$ and reference $\mathbf{x}_4$ that have different values with label $l_2$, MLSOL does not assign $l_2$ to it, because $\mathbf{s}_3$ is closer to $\mathbf{x}_4$ with Euclidean distance. In contrast, LSDMLO considers $\mathbf{s}_3$ to be associated with $l_2$, since its closest instance is $\mathbf{x}_s$ assessed by $l_2$-specific distance. Figure l2-specific distance shows that $\mathbf{s}_3$ is located in the region of $\mathcal{K}^2_{\mathbf{x}_s}$ (blue rectangle). 
Therefore, the label assignment of LSDMLO is more effective than MLSOL to guarantee label consistency in the important feature subspace.

\section{Supplementary Experiment Setup and Results}

\subsection{Hyper-Parameter Settings \label{app_sec:parameter}}

All multi-label sampling methods and classification algorithms adopt parameter configurations recommended in the respective literature.   
In heuristic methods MLSMOTE\textsubscript{U}, MLSMOTE\textsubscript{R}, RHwRSMT, MLSOL, MLBOTE and DR-SMOTE, the number of neighbors ($k$) is set to 5. 
MLSMOTE\textsubscript{U} and MLSMOTE\textsubscript{R} employ ranking and union label assignment strategies respectively.
For a fair comparison, the sampling ratio in MLROS, MLSOL, MLONC and AEMLO is set in the same way as in LSDMLO. In MLRUS, the sampling ratio is chosen in $\{ 0.1, 0.15, \dots, 0.3\}$ for small and medium datasets ($n \le 5000$), and is selected from $\{0.01,0.03, \cdots, 0.09\}$ for large datasets ($n>5000$). 
In MLBOTE, the self borderline sampling ratio $\alpha_{s}$ is searched in $\{0.01,0.02, 0.04,0.06,0.08,0.1\}$, cross-borderline selection threshold $th_{c}=7$, the number of neighbors for borderline identification $k_{b}=3$, the number of neighbors for borderline and reference selection $k_{w}=5$, and the number of loops in resampling $n_{lps}=10$.
For AEMLO, the reconstruction loss weight parameters $\alpha$ and $\beta$ are searched in $\{2^{-4}, 2^{-3}, \ldots, 2^{4}\}$.

The base binary and multi-class classifier in BR, RAkEL, CC and COCOA is the C4.5 decision tree. In MLkNN, the number of neighbors is 10. In RAkEL, the size of label subset is 3, and the number of ensemble subsets is $2q$. The number of coupled labels in COCOA is set as $\min(q-1,10)$. In C2AE, the learning rate is $10^{-5}$, the batch size is 32, and the number of epochs is 100.

\subsection{Comparison with Learning Label-Specific Features \label{app_sec:LLSF}}
Similar to learning label-specific features models \cite{LLSF}, the optimal feature weight matrix $\mathbf{W}$ obtained in Appendix Algorithm \ref{al:W} not only indicates the discrimination of features to each label but also can be employed as a multi-label classifier to predict an unseen instance $\mathbf{x}$ as $\mathbf{\hat{y}}=\mathbf{W}\mathbf{x}$. We denote the $\mathbf{W}$-based classifier as LLSF for simplicity, and its performance is obtained using the same experiment setup described in Section IV-B. 
As shown in Appendix Figure \ref{fig:LLSF_results}, compared with LLSF, LSDMLO with the two multi-label base learners achieve higher Macro-F results in most cases, indicating the effectiveness of the combination label-specific distance and multi-label oversampling. LSDMLO with COCOA performs better for the smaller dataset (scene), while LSDMLO with C2AE is more effective on larger scale datasets (Corel5k and yahoo-Business1). This shows the flexibility of sampling methods that can be incorporated with the proper multi-label base learner according to the specific characteristics of the dataset. 

\begin{figure*}[h]
\centering
\includegraphics[width=\textwidth]{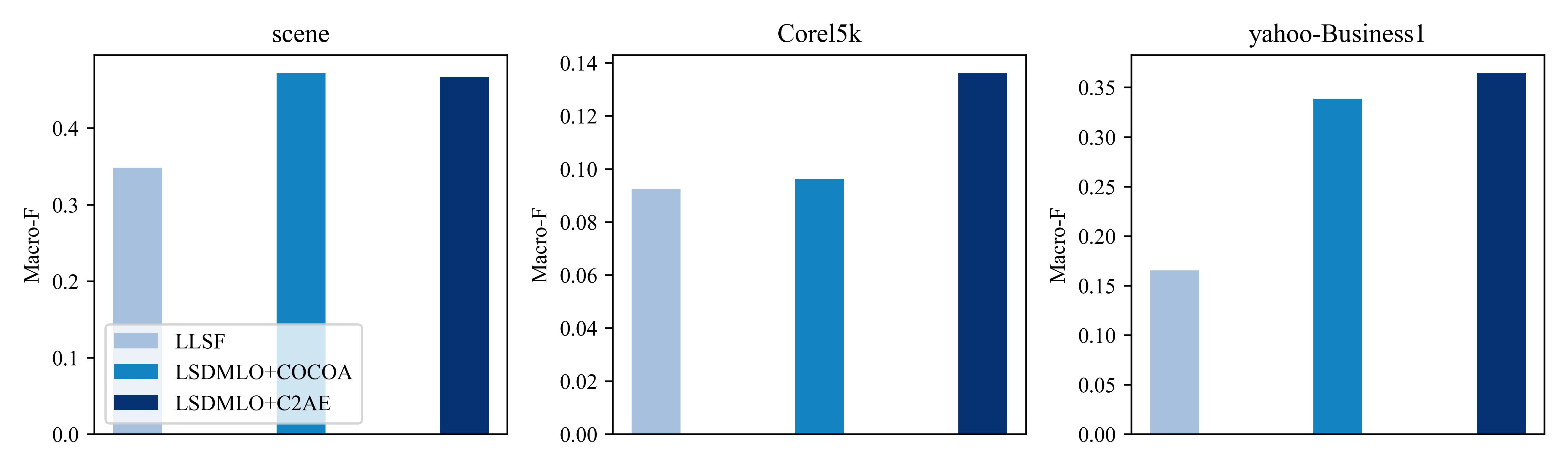} 
\caption{The Macro-F results of LLSF, LSDMO+COCOA and LSDMO+CA2E.}
\label{fig:LLSF_results}
\end{figure*}

\subsection{Parameter Analysis \label{app_sec:para_analy}}

\begin{figure*}[t]
\centering
\begin{subfigure}{0.48\textwidth}
\includegraphics[width=\textwidth]{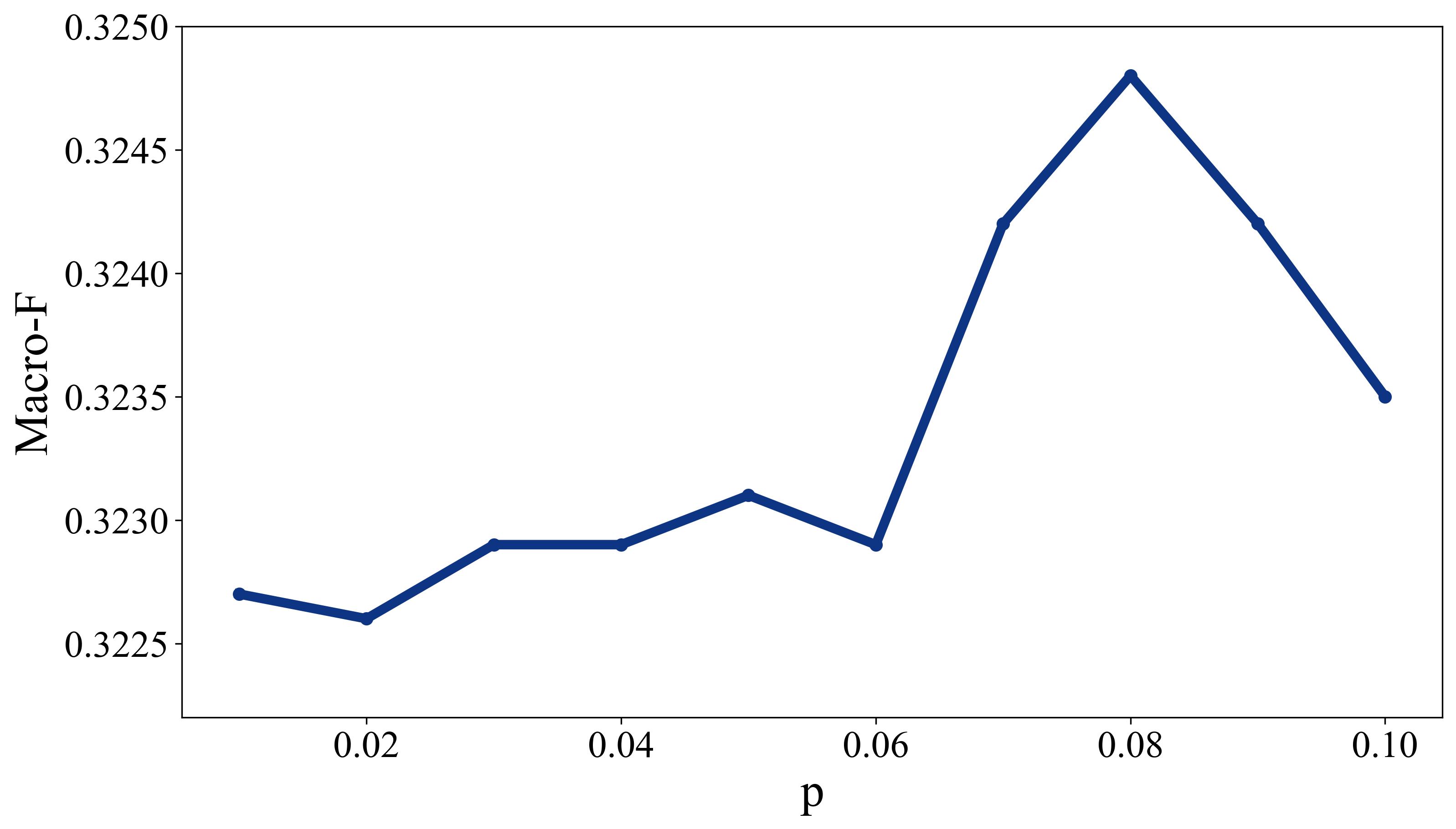} 
\caption{rcv1subset1}
\end{subfigure}
\begin{subfigure}{0.48\textwidth}
\includegraphics[width=\textwidth]{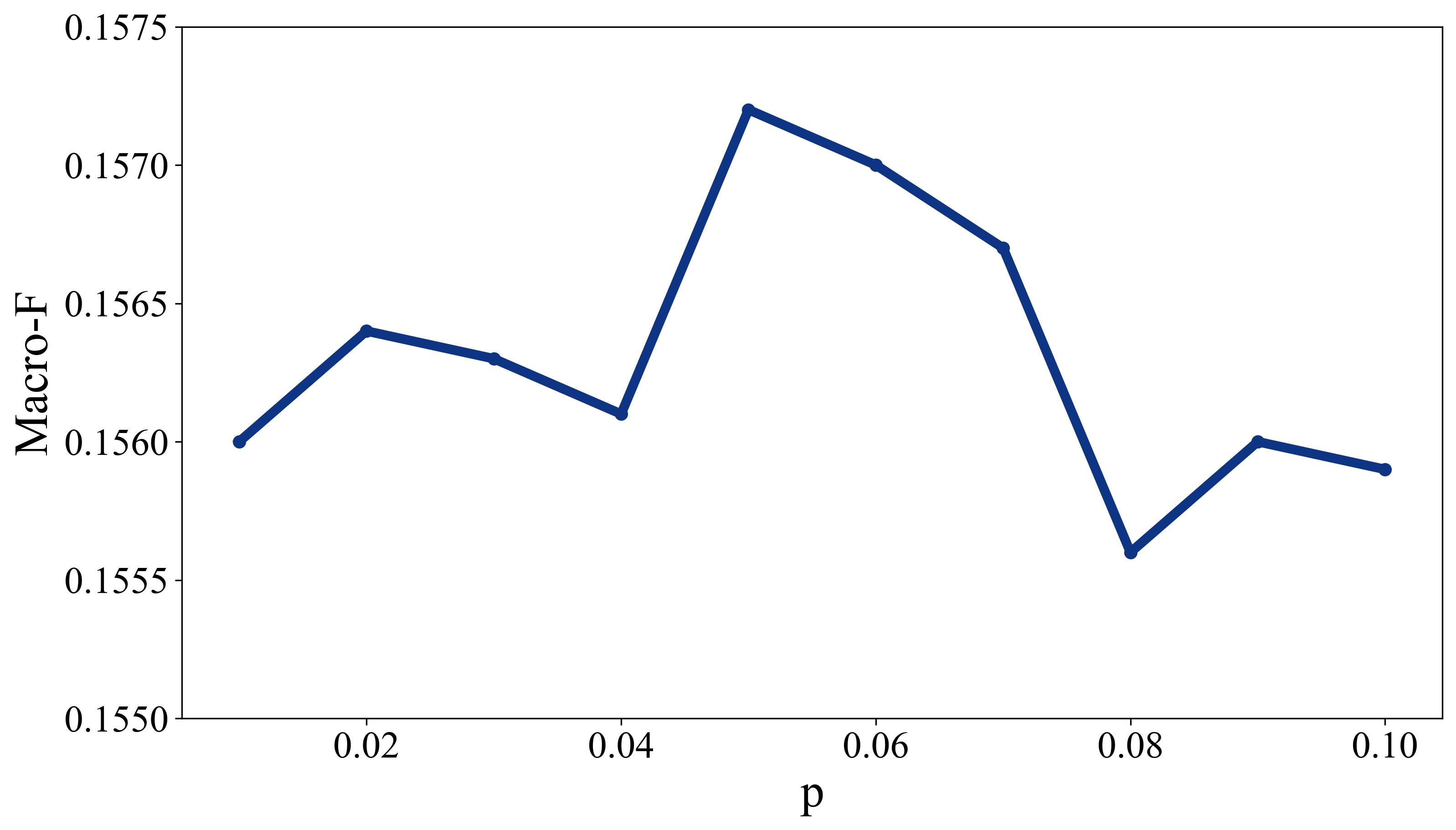}
\caption{Corel5k}
\end{subfigure}
\caption{The Macro-F results of LSDMLO with COCOA on rcv1subset1 and Corel5k datasets under various $p$.}
\label{fig:p}
\end{figure*}

Here we analyze the sensitivity of sampling ratio $p$.
As shown in Figure \ref{fig:p}, 
The performance of LSDMLO improves as $p$ increases, 
which demonstrates the effectiveness of augmenting useful synthetic instances on addressing the imbalance issue in multi-label data.
However, when $p$ becomes too large, the excessive generation of synthetic instances may alter the original distribution of the data, leading to the performance decline of LSDMLO. In general, setting $p$ around 0.06-0.08 is an acceptable choice for larger datasets. 

\begin{figure*}[!t]
\captionsetup[subfigure]{justification=centering}
\setlength{\tabcolsep}{2pt} 

\begin{subfigure}{0.3\textwidth}
 \includegraphics[width=\textwidth]{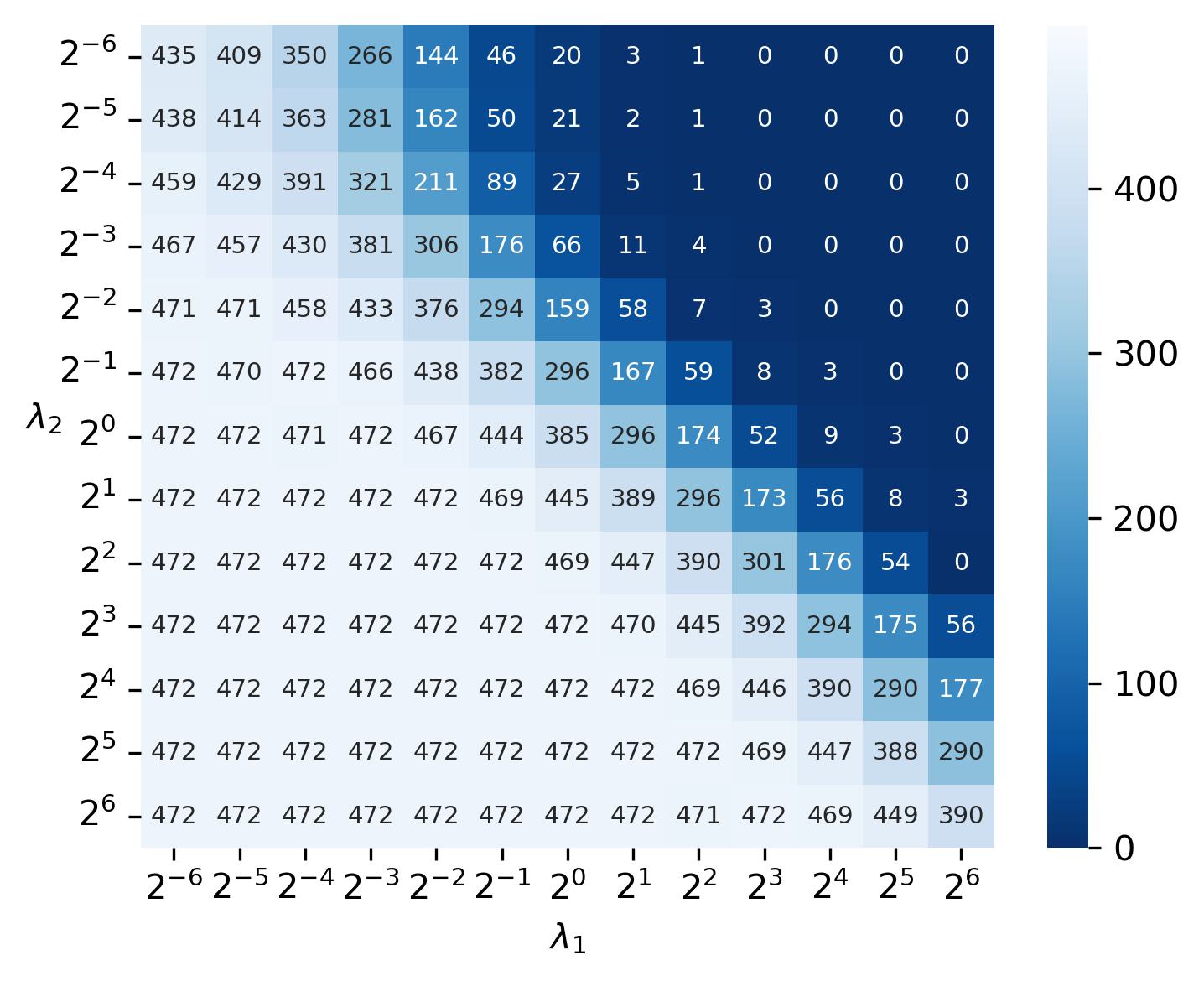}
\caption{rcvsubset1: $l_{1}$}   
\end{subfigure}
\hfill 
\begin{subfigure}{0.3\textwidth}
\includegraphics[width=\textwidth]{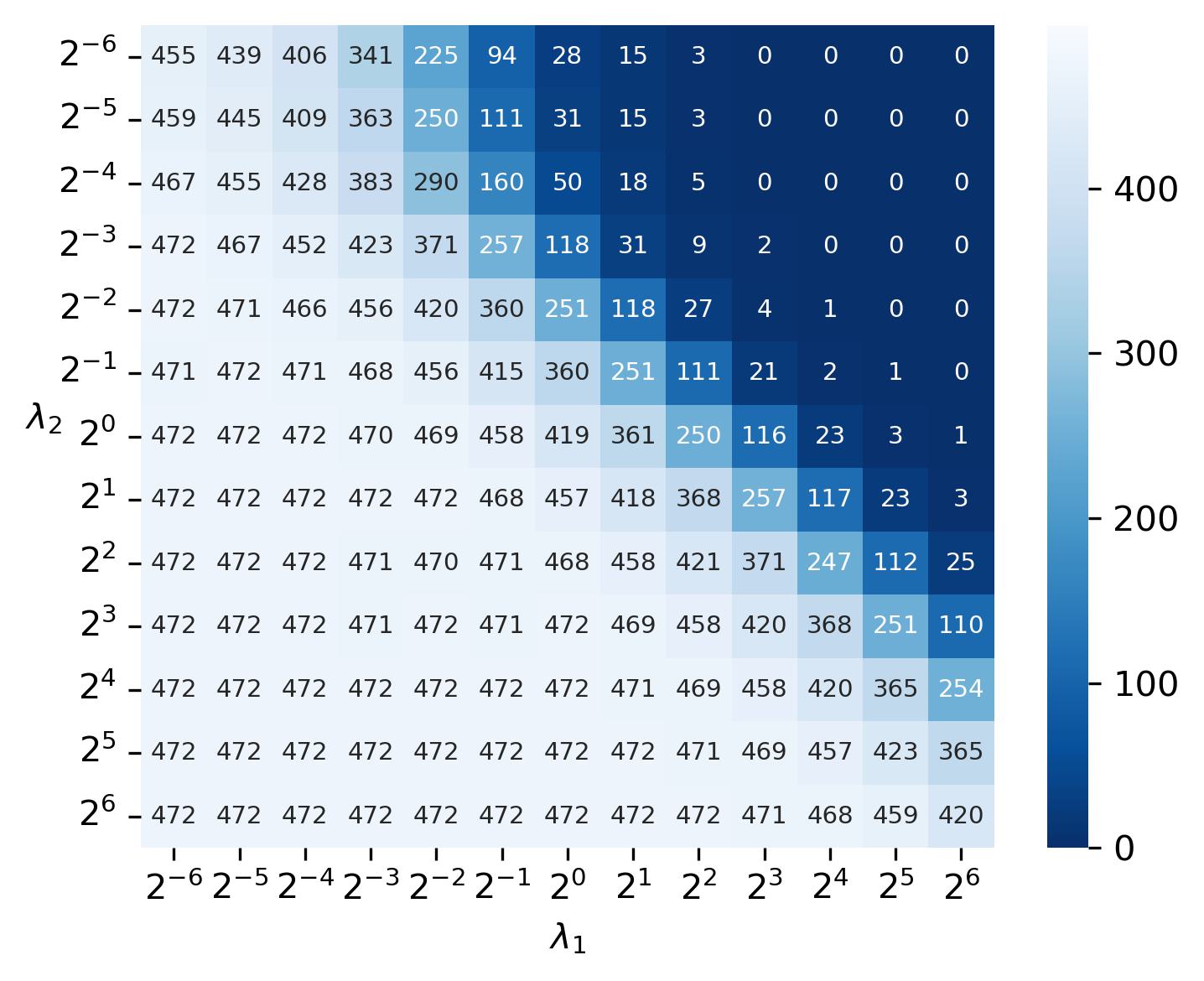}
\caption{rcvsubset1: $l_{2}$}
\end{subfigure}
\hfill
\begin{subfigure}{0.3\textwidth}
\includegraphics[width=\textwidth]{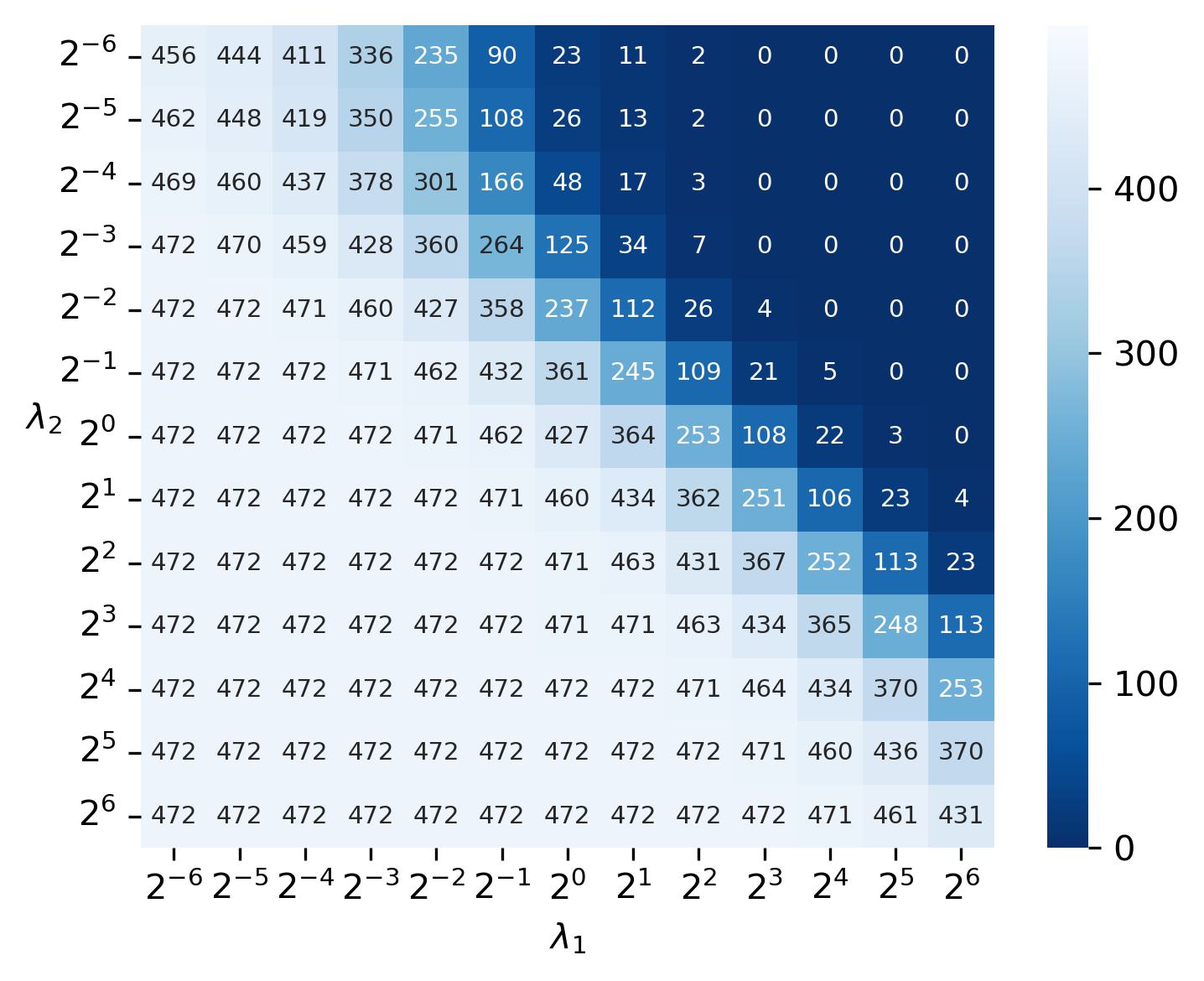}
\caption{rcvsubset1: $l_{3}$}
\end{subfigure}
\begin{subfigure}{0.3\textwidth}
\includegraphics[width=\textwidth]{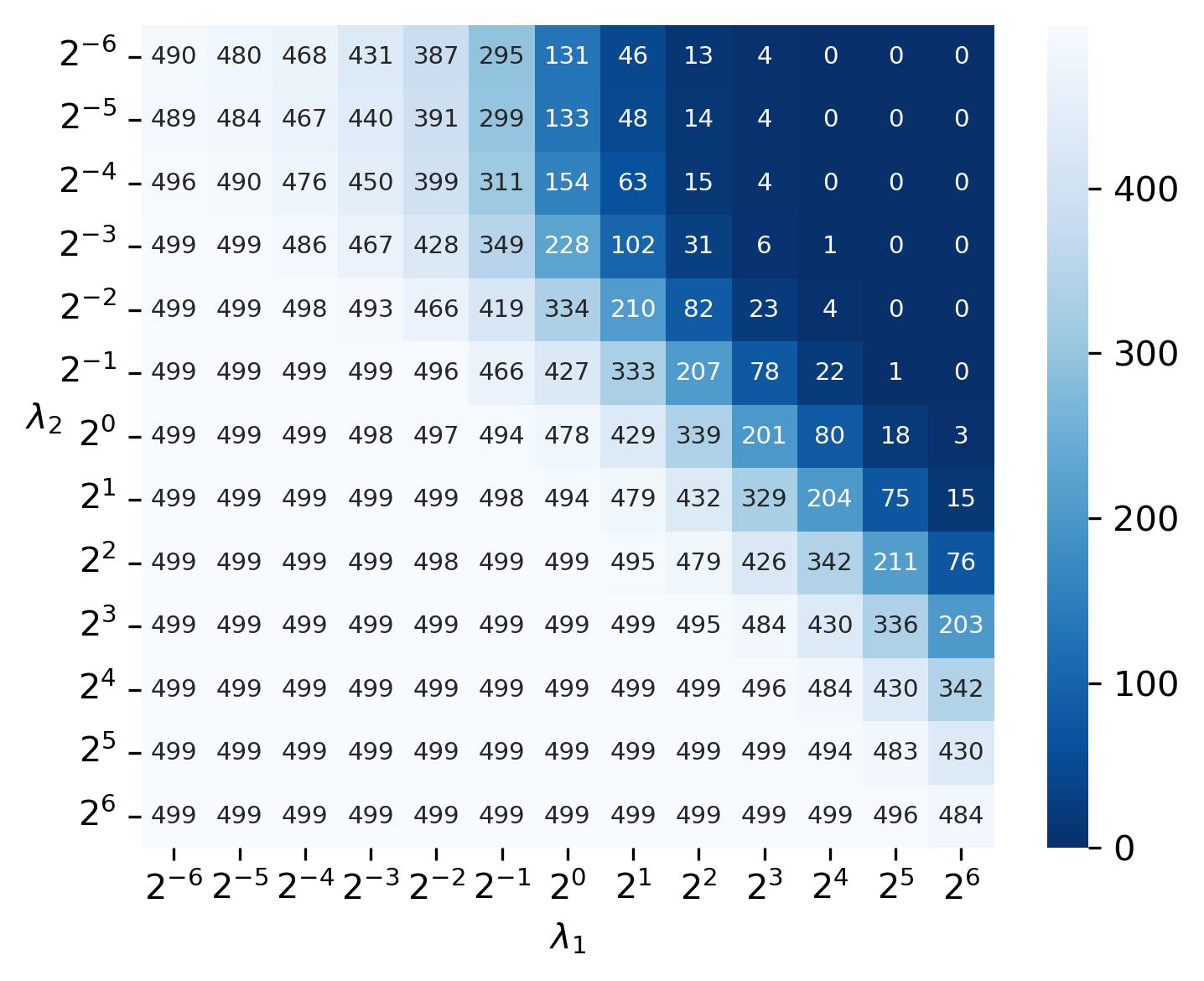}
\caption{Corel5k: $l_{1}$}
\end{subfigure}
\hfill 
\begin{subfigure}{0.3\textwidth}
\includegraphics[width=\textwidth]{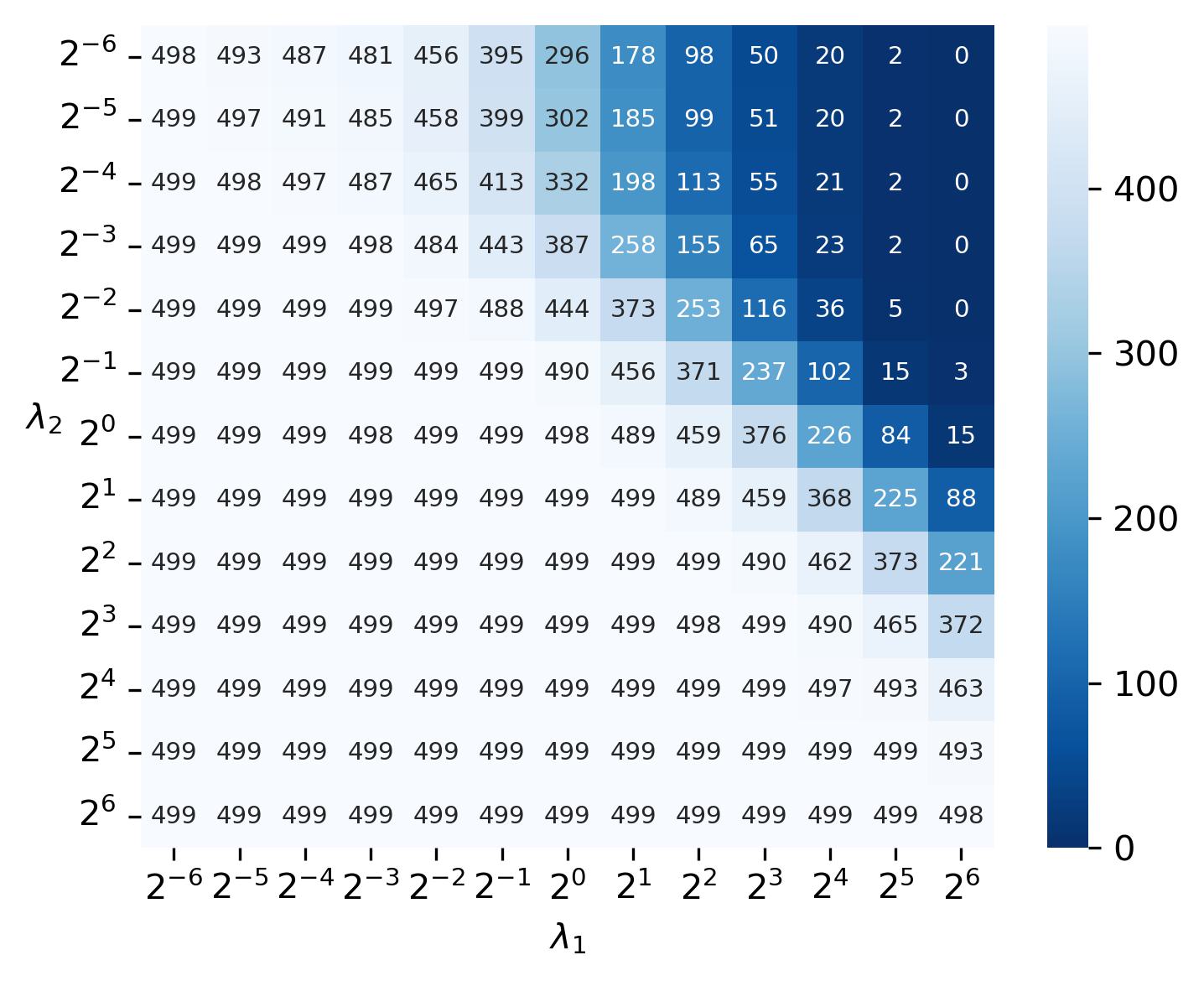}
\caption{Corel5k: $l_{2}$}
\end{subfigure}
\hfill
\begin{subfigure}{0.3\textwidth}
\includegraphics[width=\textwidth]{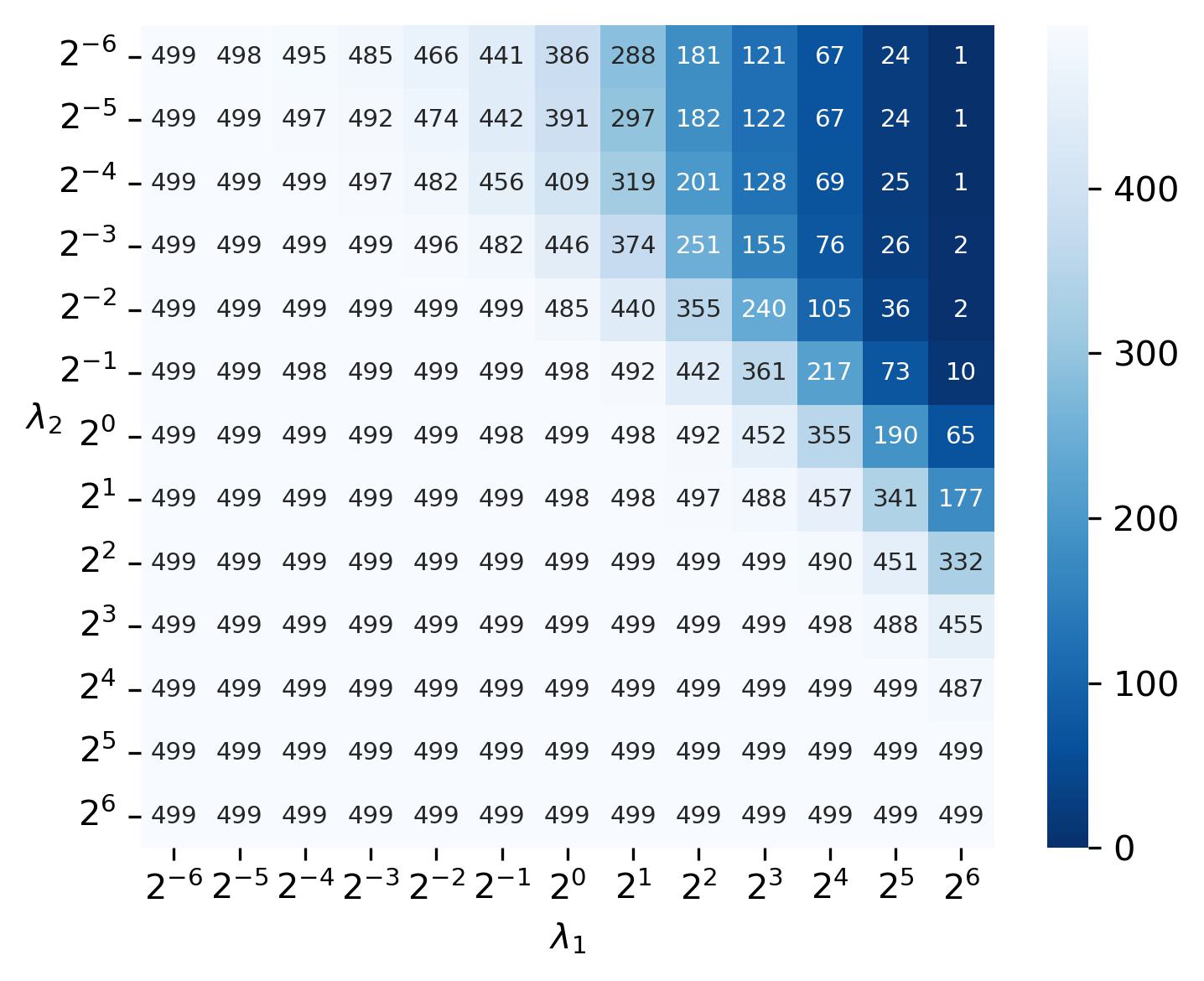}
\caption{Corel5k: $l_{3}$}
\end{subfigure}

\caption{The feature sparsity (number of features used for label-specific distance calculation) under different $\lambda_{1}$ and $\lambda_{2}$ values. The lower number in matrices implies fewer prominent features for the corresponding label.}
\label{fig:lambda}
\end{figure*}

Trade-off parameters $\lambda_{1}$ and $\lambda_{2}$ impact the sparsity of $\mathbf{W}$. Because only features with non-zero weights contribute to label-specific distance computation, the number of non-zero elements in row $\mathbf{w}_{\cdot j}$ is equal to the number of relevant features for label $l_j$.
Figure \ref{fig:lambda} illustrates the influence of $\lambda_{1}$ and $\lambda_{2}$ on feature sparsity, i.e., the number of features used for label-specific distance calculation.  
An ascending trend in $\lambda_{1}$ stresses on $l_1$ norm of the weight matrix, leading to more sparse features. 
On the other hand, larger $\lambda_{2}$ emphasizes the label correlation regularizer, mitigating the sparsity of features.


\begin{figure*}[h]
\centering 
\begin{subfigure}{0.45\textwidth}
  \includegraphics[width=\textwidth]{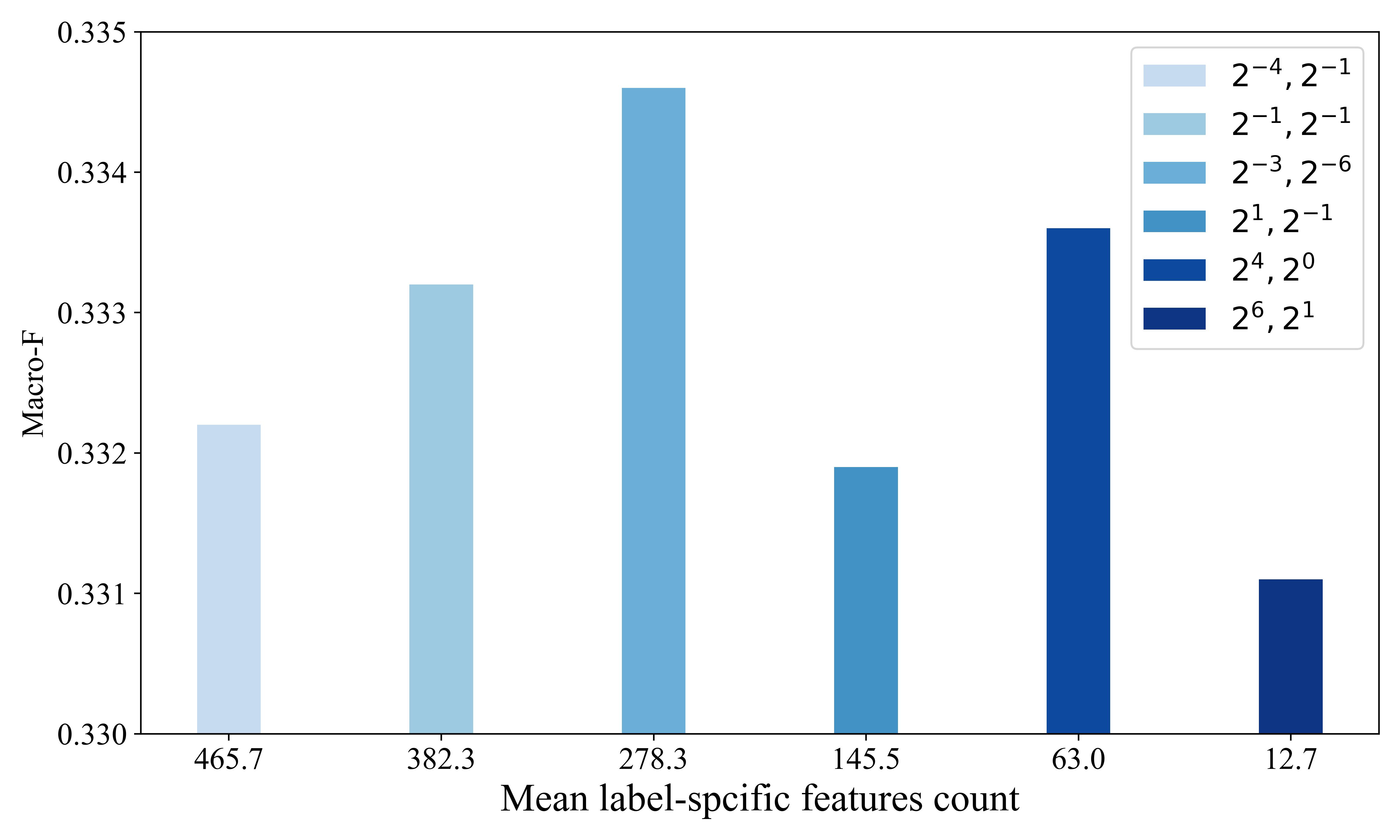}
  \caption{rcv1subset1}
\end{subfigure}
\begin{subfigure}{0.45\textwidth}
\includegraphics[width=\textwidth]{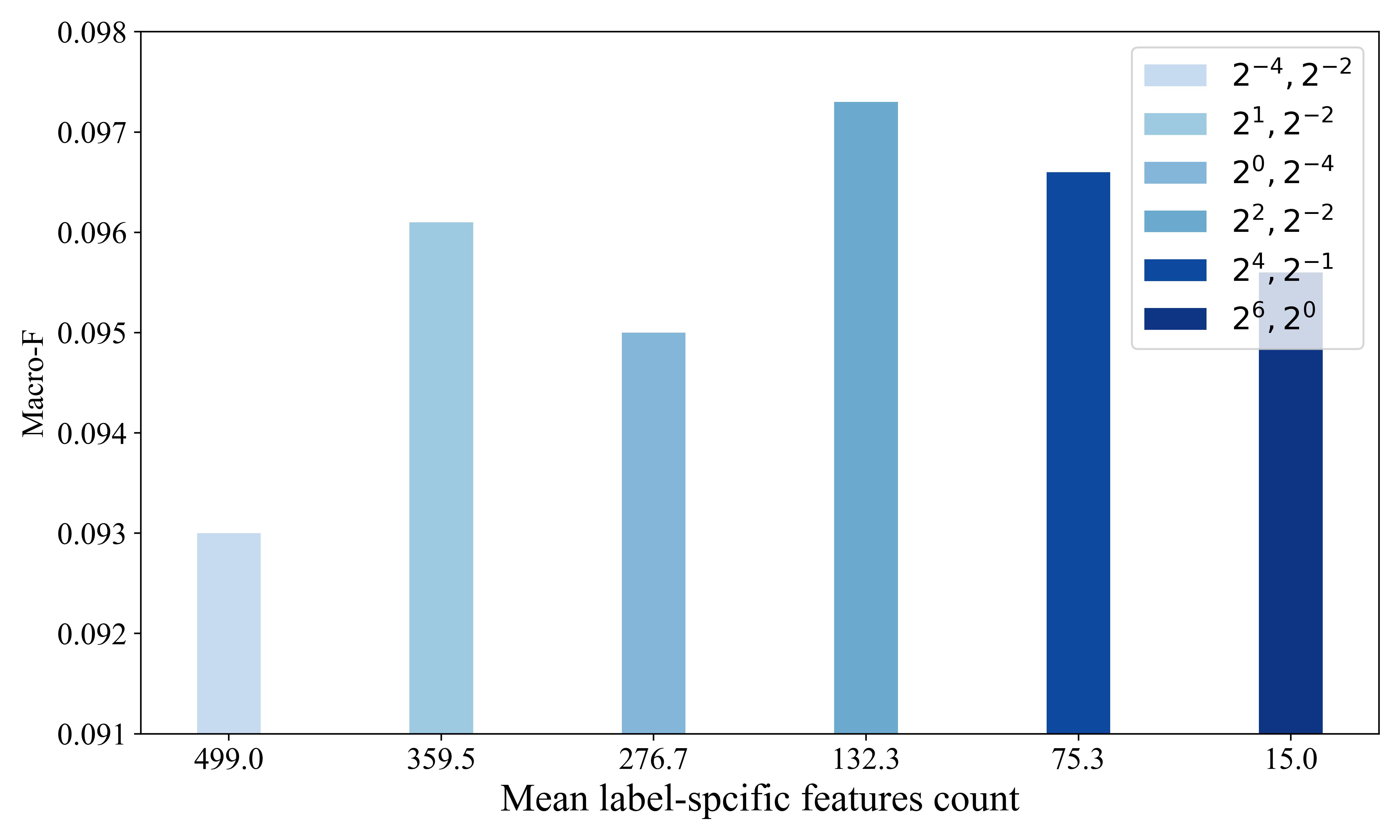}
\caption{Corel5k}
\end{subfigure}
\caption{The Macro-F results of LSDMLO with COCOA under various levels of feature sparsity ($\lambda_1$ and $\lambda_2$ combinations) on the rcv1subset1 and Corel5k datasets. Feature dimensions of rcv1subset1 and Corelk5 datasets are 499 and 472, respectively.}
\label{fig:F}
\end{figure*}


Furthermore, we analyze the influence of feature sparsity on the performance of LSDMLO combined with COCOA on rcv1subset1 and Corel5k datasets.
As shown in Figure \ref{fig:F}, the performance increases as the number of label-specific features decreases. The sparser features exclude the impact of irrelevant features, thereby projecting into a more accurate subspace to measure distances between examples.
The peak result approaches when  56\% and 28\% prominent features are retained for label-specific distance calculation on rcv1subset1 and Corel5k datasets, respectively.
The performance starts to deteriorate when features become extremely sparse, because filtering more features would overlook potentially crucial information.



\subsection{Sampling Time Comparison\label{app_sec:time}}



\begin{figure*}[h]
\centering 
\begin{subfigure}{0.45\textwidth}
 \includegraphics[width=\textwidth]{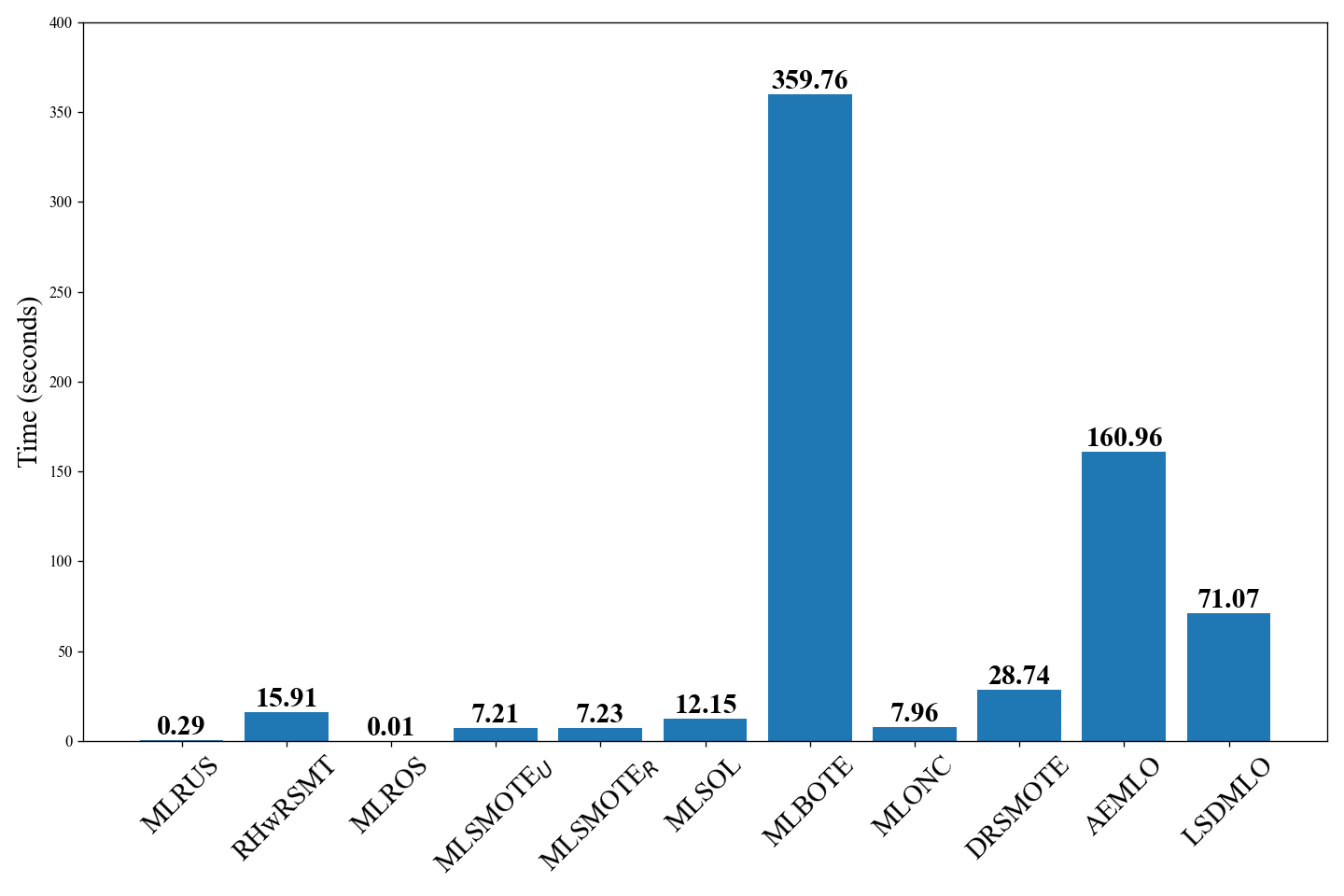}
\caption{rcv1subset1}   
\end{subfigure}
\begin{subfigure}{0.45\textwidth}
 \includegraphics[width=\textwidth]{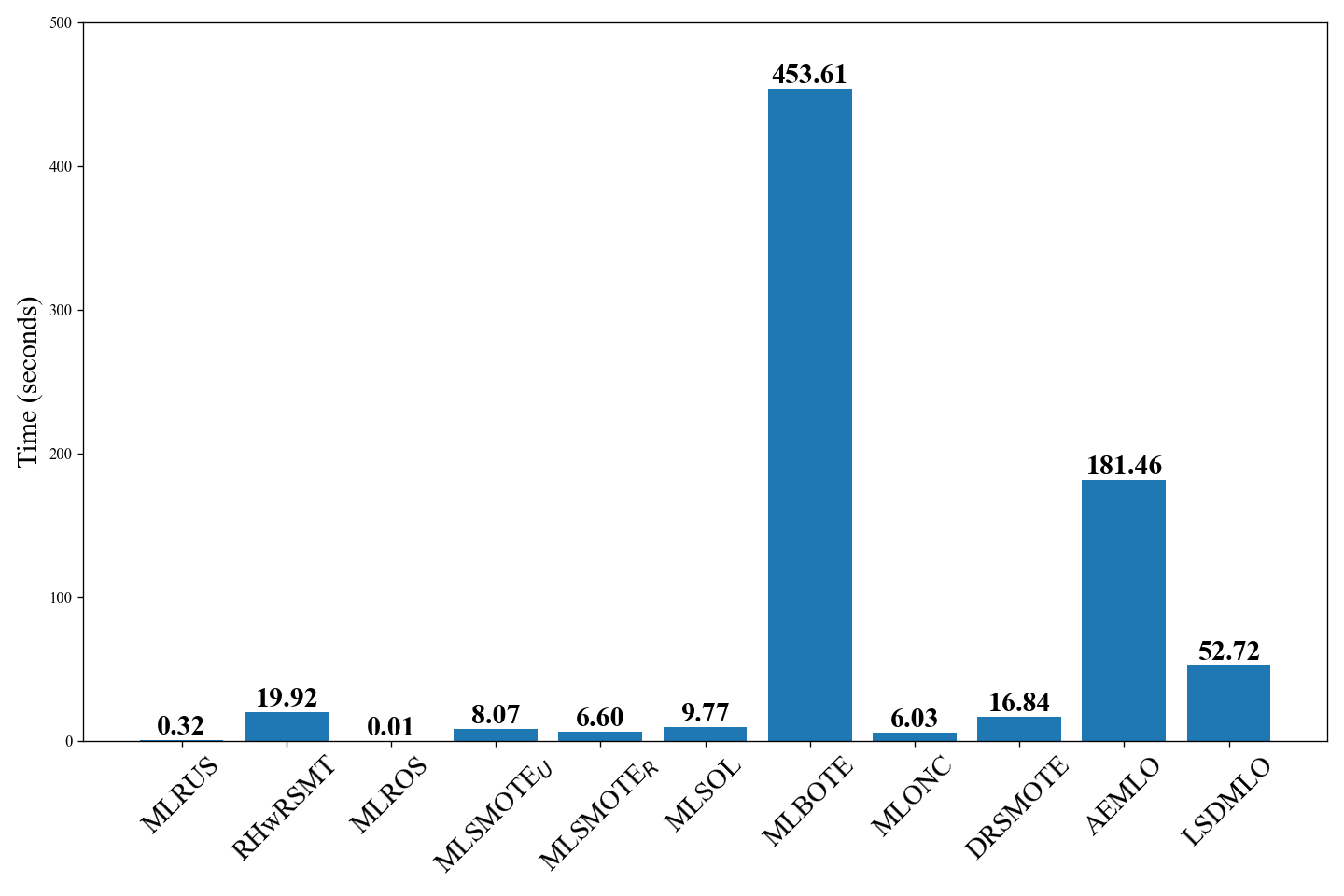}
\caption{Corel5k}   
\end{subfigure}

\caption{The sampling time of different multi-label sampling methods on the rcv1subset1 and Corel5k datasets.}
\label{fig:samplingtime}
\end{figure*}

The sampling times of multi-label sampling methods on rcv1subset1 and Corel5k datasets are shown in Figure \ref{fig:samplingtime}.
To ensure a fair comparison, each oversampling method generates  \(0.1 \times n\) synthetic instances, and MLRUS deletes \(0.1 \times n\) examples. 

MLROS and MLRUS are the two fastest methods, duplicating and removing instances based on label statistics. 
Two variants of MLSMOTE and MLONC, which involve searching for $k$NNs from each minority instance set, exhibit slower sampling procedures than MLROS and MLRUS. 
In contrast to MLSMOTE, MLSOL necessitates retrieving the nearest neighbor from the entire dataset, resulting in a higher time cost. 
DR-SMOTE requires identifying reference samples with different label sets for each seed instance, which introduces additional distance calculations and results in higher sampling time than MLSMOTE and MLSOL.
RHwRSMT, which decouples some high-scramble instances, consumes more time than MLSMOTE. 
Most sampling time consumed by LSDMLO is dedicated to acquiring $\mathbf{W}$ and measuring label-specific nearest neighbors. Although label-specific distance involves calculating partial features, the iterative APG algorithm makes LSDMLO slower than most neighbor-based sampling approaches.  
AEMLO requires training a deep autoencoder to learn latent representations before generating synthetic instances, which introduces a substantial computational overhead and makes it significantly much slower than LSDMLO.
MLBOTE is the most time-consuming method, as it computes the neighborhood for each label when determining the instance type.

\subsection{Detailed Results of Sampling Methods}

\begin{table*}[h]
\caption{Results of sampling methods with BR base learner}
\centering
\resizebox{1.0\textwidth}{!}{
\begin{tabular}{ccccccccccccccc}
\toprule
& Datasets & Base & MLRUS & RHwRSMT & MLROS & MLSMOTE\textsubscript{U} & MLSMOTE\textsubscript{R} & MLSOL & MLBOTE & MLONC & DR-SMOTE & AEMLO & LSDMLO  \\  \midrule 
\multirow{14}{*}{Macro-F} 
& emotion & 0.5571(11) & 0.5574(10) & 0.3915(12) & 0.5626(7) & 0.5589(8) & 0.5641(5) & 0.5626(6) & \textbf{0.5692(1)} & 0.5642(4) & 0.5575(9) & 0.5660(3) & 0.5671(2) \\
& scene & 0.5850(8) & 0.5893(4) & 0.5684(12) & 0.5855(6) & 0.5815(9) & 0.5805(10) & 0.5857(5) & 0.5855(7) & 0.5896(3) & 0.5791(11) & \textbf{0.5958(1)} & 0.5940(2) \\
& yeast & 0.3883(9) & 0.3887(7) & 0.2741(12) & 0.3832(11) & 0.3832(10) & 0.3885(8) & 0.3944(4) & 0.3945(3) & 0.3895(6) & \textbf{0.3972(1)} & 0.3941(5) & 0.3961(2) \\
& Corel5k & 0.0591(9) & 0.0572(10) & 0.0417(12) & 0.0601(4) & 0.0597(5) & 0.0596(6) & 0.0592(8) & 0.0595(7) & \textbf{0.0610(1)} & 0.0532(11) & 0.0606(3) & 0.0607(2) \\
& rcv1 & 0.2275(5) & 0.2248(6) & 0.1162(12) & 0.2239(8) & 0.2222(11) & 0.2237(9) & 0.2243(7) & 0.2284(3) & 0.2277(4) & 0.2232(10) & \textbf{0.2296(1)} & 0.2292(2) \\
& rcv2 & 0.2011(6) & 0.1995(8) & 0.1084(12) & 0.2016(5) & 0.1974(11) & 0.1983(10) & 0.2031(2) & 0.2008(7) & 0.2023(4) & 0.1987(9) & 0.2025(3) & \textbf{0.2033(1)} \\
& rcv3 & 0.2079(4) & 0.2052(9) & 0.1128(12) & 0.2055(8) & 0.2011(10) & 0.2079(5) & 0.2087(2) & 0.2076(7) & 0.2077(6) & 0.1984(11) & \textbf{0.2109(1)} & 0.2084(3) \\
\bottomrule
\end{tabular}
}
\label{tab:BR}
\end{table*}
\clearpage

\begin{table*}[!h]
\caption{Results of sampling methods with MLkNN base learner}
\centering
\resizebox{1.0\textwidth}{!}{
\begin{tabular}{cccccccccccccc}
\toprule
& Datasets & Base & MLRUS & RHwRSMT & MLROS & MLSMOTE\textsubscript{U} &  MLSMOTE\textsubscript{R} &  MLSOL & MLBOTE & MLONC & DR-SMOTE & AEMLO & LSDMLO  \\  \midrule 
\multirow{14}{*}{Macro-F} 
& emotion & 0.4313(9) & 0.4216(11) & 0.2630(12) & 0.4325(8) & 0.4325(7) & 0.4944(2) & 0.4885(3) & \textbf{0.5295(1)} & 0.4552(5) & 0.4243(10) & 0.4356(6) & 0.4787(4) \\
& scene & 0.7348(5) & 0.7235(12) & 0.7298(10) & 0.7320(9) & 0.7320(8) & 0.7433(3) & 0.7293(11) & 0.7441(2) & \textbf{0.7499(1)} & 0.7326(7) & 0.7357(4) & 0.7340(6) \\
& yeast & 0.4006(10) & 0.4035(9) & 0.2875(12) & 0.4051(8) & 0.4051(7) & 0.4131(5) & \textbf{0.4358(1)} & 0.4122(6) & 0.4158(3) & 0.4341(2) & 0.3921(11) & 0.4158(4) \\
& Corel5k & 0.0344(6) & 0.0349(4) & 0.0154(12) & \textbf{0.0375(1)} & 0.0262(10) & 0.0344(7) & 0.0324(8) & 0.0351(2) & 0.0317(9) & 0.0238(11) & 0.0351(3) & 0.0346(5) \\
& rcv1 & 0.1376(10) & 0.1361(11) & 0.0973(12) & 0.1673(9) & 0.1673(8) & 0.1847(3) & \textbf{0.2242(1)} & 0.1795(4) & 0.1778(5) & 0.1708(7) & 0.1714(6) & 0.1937(2) \\
& rcv2 & 0.1375(10) & 0.1353(11) & 0.0856(12) & 0.1575(9) & 0.1575(8) & 0.1745(5) & \textbf{0.2063(1)} & 0.1612(7) & 0.1815(3) & 0.1682(6) & 0.1754(4) & 0.1887(2) \\
& rcv3 & 0.1331(10) & 0.1316(11) & 0.0801(12) & 0.1582(8) & 0.1580(9) & 0.1732(3) & \textbf{0.1988(1)} & 0.1596(7) & 0.1675(6) & 0.1688(5) & 0.1702(4) & 0.1954(2) \\
\bottomrule
\end{tabular}
}
\label{tab:MLkNN}
\end{table*}
\clearpage

\begin{table*}[!h]
\centering
\caption{Results of sampling methods with CC base learner}
\resizebox{1.0\textwidth}{!}{
\begin{tabular}{ccccccccccccccc}
\toprule
& Datasets & Base & MLRUS & RHwRSMT & MLROS & MLSMOTE\textsubscript{U} & MLSMOTE\textsubscript{R} & MLSOL & MLBOTE & MLONC & DR-SMOTE & AEMLO & LSDMLO  \\  \midrule 
\multirow{14}{*}{Macro-F} 
& emotion & 0.5549(8) & 0.5540(11) & 0.4478(12) & 0.5546(9) & 0.5602(6) & 0.5615(5) & 0.5583(7) & 0.5546(10) & 0.5713(2) & 0.5630(4) & \textbf{0.5755(1)} & 0.5638(3) \\
& scene & 0.5865(6) & 0.5848(7) & 0.5611(12) & 0.5887(4) & 0.5757(10) & 0.5737(11) & 0.5830(8) & 0.5887(5) & 0.5895(3) & 0.5812(9) & \textbf{0.5930(1)} & 0.5903(2) \\
& yeast & 0.3826(10) & 0.3913(4) & 0.3036(12) & 0.3826(11) & 0.3879(7) & 0.3858(9) & 0.3906(6) & 0.3863(8) & 0.3918(3) & 0.3910(5) & 0.3924(2) & \textbf{0.3977(1)} \\
& Corel5k & 0.0509(11) & 0.0568(6) & 0.0412(12) & 0.0577(4) & 0.0577(3) & 0.0555(9) & 0.0564(8) & 0.0578(2) & \textbf{0.0615(1)} & 0.0512(10) & 0.0566(7) & 0.0576(5) \\
& rcv1 & 0.2182(9) & 0.2173(10) & 0.1442(12) & 0.2185(6) & 0.2152(11) & 0.2203(5) & 0.2217(3) & 0.2185(7) & \textbf{0.2388(1)} & 0.2183(8) & 0.2215(4) & 0.2223(2) \\
& rcv2 & 0.1967(10) & 0.1985(8) & 0.1362(12) & 0.2003(4) & 0.1926(11) & 0.1991(7) & 0.2008(3) & 0.1996(6) & \textbf{0.2241(1)} & 0.1973(9) & 0.2030(2) & 0.2002(5) \\
& rcv3 & 0.2017(9) & 0.2009(10) & 0.1345(12) & 0.2046(5) & 0.1972(11) & 0.2022(8) & 0.2052(4) & 0.2032(6) & \textbf{0.2236(1)} & 0.2027(7) & 0.2060(2) & 0.2054(3) \\
\bottomrule
\end{tabular}
}
\label{tab:CC}
\end{table*}
\clearpage

\begin{table*}[!h]
\caption{Results of sampling methods with RAkEL base learner}
\centering
\resizebox{1.0\textwidth}{!}{
\begin{tabular}{cccccccccccccc}
\toprule
& Datasets & Base & MLRUS & RHwRSMT & MLROS & MLSMOTE\textsubscript{U} & MLSMOTE\textsubscript{R} & MLSOL & MLBOTE & MLONC & DR-SMOTE & AEMLO & LSDMLO  \\  \midrule 
\multirow{14}{*}{Macro-F} 
& emotion & 0.5458(11) & 0.5465(10) & 0.4328(12) & 0.5568(6) & 0.5614(2) & 0.5571(5) & 0.5549(7) & 0.5546(8) & \textbf{0.5645(1)} & 0.5511(9) & 0.5577(4) & 0.5579(3) \\
& scene & 0.5807(9) & 0.5842(4) & 0.5699(12) & 0.5834(5) & 0.5797(11) & 0.5812(8) & 0.5821(7) & 0.5828(6) & 0.5855(3) & 0.5803(10) & \textbf{0.5949(1)} & 0.5866(2) \\
& yeast & 0.3803(9) & 0.3789(10) & 0.3034(12) & 0.3855(5) & 0.3823(8) & 0.3832(7) & 0.3925(2) & 0.3841(6) & 0.3770(11) & 0.3908(4) & 0.3918(3) & \textbf{0.3926(1)} \\
& Corel5k & 0.0566(10) & 0.0575(9) & 0.0411(12) & 0.0585(6) & 0.0581(8) & 0.0559(11) & 0.0603(2) & 0.0589(4) & 0.0595(3) & 0.0585(7) & 0.0587(5) & \textbf{0.0604(1)} \\
& rcv1 & 0.2139(9) & 0.2155(8) & 0.1236(12) & 0.2159(7) & 0.2138(10) & 0.2136(11) & 0.2174(5) & 0.2165(6) & 0.2203(2) & 0.2178(4) & 0.2190(3) & \textbf{0.2206(1)} \\
& rcv2 & 0.1924(10) & 0.1947(8) & 0.1142(12) & 0.1987(4) & 0.1905(11) & 0.1975(6) & 0.1929(9) & 0.1983(5) & 0.1995(3) & 0.1961(7) & 0.1997(2) & \textbf{0.1999(1)} \\
& rcv3 & 0.2009(9) & 0.1999(10) & 0.1166(12) & 0.2013(8) & 0.1968(11) & 0.2032(7) & 0.2044(2) & 0.2044(3) & 0.2036(5) & 0.2034(6) & 0.2042(4) & \textbf{0.2048(1)} \\
\bottomrule
\end{tabular}
}
\label{tab:RAkEL}
\end{table*}
\clearpage

\begin{table*}[!h]
\caption{Results of sampling methods with COCOA base learner}
\centering
\resizebox{1.0\textwidth}{!}{
\begin{tabular}{cccccccccccccc}
\toprule
& Datasets & Base & MLRUS & RHwRSMT & MLROS & MLSMOTE\textsubscript{U} & MLSMOTE\textsubscript{R} & MLSOL & MLBOTE & MLONC & DR-SMOTE & AEMLO & LSDMLO  \\  \midrule 
\multirow{14}{*}{Macro-F} 
 & emotion & 0.6239(11) & 0.6267(10) & 0.5773(12) & 0.6277(9) & 0.6277(8) & 0.6344(6) & 0.6306(7) & 0.6360(2) & 0.6351(5) & 0.6352(4) & 0.6357(3) & \textbf{0.6361(1)} \\
 & scene & 0.7214(3) & 0.7197(9) & 0.7173(12) & 0.7203(7) & 0.7194(11) & 0.7200(8) & 0.7207(5) & 0.7209(4) & \textbf{0.7314(1)} & 0.7196(10) & 0.7206(6) & 0.7216(2) \\
 & yeast & 0.4565(11) & 0.4711(4) & 0.4679(10) & 0.4679(9) & 0.4689(6) & 0.4714(3) & 0.4689(7) & 0.4719(2) & 0.4564(12) & 0.4686(8) & 0.4697(5) & \textbf{0.4721(1)} \\
 & Corel5k & 0.0957(5) & 0.0921(11) & 0.0875(12) & \textbf{0.0963(1)} & 0.0955(7) & 0.0942(10) & 0.0961(3) & 0.0957(6) & 0.0952(9) & 0.0960(4) & 0.0954(8) & 0.0962(2) \\
 & rcv1 & 0.2969(11) & 0.2960(12) & 0.3002(10) & 0.3263(4) & 0.3257(6) & 0.3257(7) & 0.3246(8) & 0.3229(9) & 0.3262(5) & 0.3278(3) & 0.3305(2) & \textbf{0.3334(1)} \\
 & rcv2 & 0.2790(12) & 0.2797(11) & 0.2879(10) & 0.3090(4) & 0.3058(8) & 0.3099(2) & 0.3088(5) & 0.3083(6) & 0.3035(9) & 0.3081(7) & 0.3093(3) & \textbf{0.3103(1)} \\
 & rcv3 & 0.2806(11) & 0.2799(12) & 0.2965(10) & \textbf{0.3140(1)} & 0.3071(8) & 0.3081(6) & 0.3089(5) & 0.2999(9) & 0.3110(2) & 0.3075(7) & 0.3095(4) & 0.3103(3) \\
\bottomrule
\end{tabular}
}
\label{tab:COCOA}
\end{table*}
\clearpage

\begin{table*}[!h]
\caption{Results of sampling methods with C2AE deep learner}
\centering
\resizebox{1.0\textwidth}{!}{
\begin{tabular}{cccccccccccccc}
\toprule
& Datasets & Base & MLRUS & RHwRSMT & MLROS & MLSMOTE\textsubscript{U} & MLSMOTE\textsubscript{R} & MLSOL & MLBOTE & MLONC & DR-SMOTE & AEMLO & LSDMLO  \\  \midrule 
\multirow{14}{*}{Macro-F} 
 & emotion & 0.5373(11) & 0.5938(4) & 0.5193(12) & 0.5651(10) & \textbf{0.6099(1)} & 0.5994(2) & 0.5962(3) & 0.5938(5) & 0.5796(9) & 0.5865(8) & 0.5902(7) & 0.5908(6) \\
 & scene & 0.7884(6) & 0.7808(10) & 0.7936(3) & 0.7823(9) & 0.7749(12) & 0.7792(11) & 0.7868(8) & 0.7898(5) & 0.7914(4) & 0.7884(7) & 0.7948(2) & \textbf{0.7971(1)} \\
 & yeast & 0.4111(11) & 0.4124(10) & 0.4084(12) & 0.4320(6) & 0.4251(7) & 0.4186(8) & 0.4146(9) & 0.4424(2) & 0.4406(4) & 0.4383(5) & 0.4419(3) & \textbf{0.4672(1)} \\
 & Corel5k & 0.1281(10) & 0.1333(4) & 0.1077(12) & 0.1294(9) & 0.1322(6) & 0.1201(11) & \textbf{0.1362(1)} & 0.1361(2) & 0.1321(7) & 0.1329(5) & 0.1318(8) & 0.1340(3) \\
 & rcv1 & 0.3205(10) & 0.3329(4) & 0.2068(12) & 0.3314(7) & 0.3160(11) & 0.3315(6) & 0.3225(9) & 0.3329(5) & 0.3363(2) & 0.3264(8) & 0.3342(3) & \textbf{0.3388(1)} \\
 & rcv2 & 0.2640(12) & 0.3191(9) & 0.2908(11) & 0.3322(2) & 0.2971(10) & 0.3264(6) & 0.3285(4) & 0.3291(3) & 0.3284(5) & 0.3227(8) & 0.3239(7) & \textbf{0.3346(1)} \\
 & rcv3 & 0.2301(12) & 0.2469(10) & 0.2450(11) & \textbf{0.2924(1)} & 0.2551(9) & 0.2657(4) & 0.2673(3) & 0.2569(8) & 0.2602(6) & 0.2578(7) & 0.2657(5) & 0.2822(2) \\
\bottomrule
\end{tabular}
}
\label{tab:C2AE}
\end{table*}
\clearpage

\bibliographystyle{splncs04}
\bibliography{ECML2026-CameraReady/ref}